\documentclass[runningheads]{llncs}
\usepackage{makeidx}
\usepackage{graphicx}
\usepackage{amsmath,amssymb} 
\usepackage{color}

\usepackage{subcaption}
\usepackage{multicol}
\usepackage{gensymb}

\usepackage{hyperref}

\newcommand\fulltitle{Achieving Generalizable Robustness of Deep Neural Networks by Stability Training}
\newcommand\shorttitle{Achieving Generalizable Robustness by Stability Training}

\begin{document}
	\pagestyle{headings}
	\mainmatter

	\def\GCPR19SubNumber{74}

	\title{\fulltitle}

	
	
\author{Jan Laermann\inst{1} \and
Wojciech Samek\inst{1}  \and
Nils Strodthoff\inst{1} }

\institute{Fraunhofer Heinrich Hertz Institute, Einsteinufer 37, 10587 Berlin, Germany\\
\email{firstname.lastname@hhi.fraunhofer.de}}
\titlerunning{\shorttitle}

\newcommand\colnst[1]{{\color{red}#1}}
\newcommand\coljl[1]{{\color{blue}#1}}
\newcommand\colws[1]{{\color{green}#1}}

\newcommand{\Lzero}{L_0}
\newcommand{\Lstab}{L_\text{stab}}
\newcommand{\Lnosym}{L_\text{nosym}}
\newcommand{\Lfsym}{L_\text{fsym}}
\newcommand{\Lpsym}{L_\text{psym}}


\maketitle

\begin{abstract}
We study the recently introduced stability training as a general-purpose method to increase the robustness of deep neural networks against input perturbations.
In particular, we explore its use as an alternative to data augmentation and validate its performance against a number of distortion types and transformations including adversarial examples.
In our image classification experiments using ImageNet data stability training performs on a par or even outperforms data augmentation for specific transformations, while consistently offering improved robustness against a broader range of distortion strengths and types unseen during training, a considerably smaller hyperparameter dependence and less potentially negative side effects compared to data augmentation.
\end{abstract}

\section{Introduction}
Deep neural networks (DNN) are  complex learning systems, which have been used in a variety
of tasks with great success in recent times. In some fields, like visual recognition or playing
games, DNNs can compete with or even outperform their human counterparts  \cite{he_delving_2015,silver_mastering_2016}, showcasing their utility and effectiveness.

In real-world applications, however, there are a number of quality criteria that go beyond single scalar 
performance metrics such as classification accuracy that are typically considered when comparing 
different classification algorithms. These quality criteria include interpretability \cite{MonDSP18}, the quantification of uncertainty \cite{gal2016uncertainty} 
but also robustness in a general sense. The aspect of robustness comprises both robustness against label noise, see \cite{frenay2014classification} for a review, 
and robustness against any kind of input noise. In this work we are only concerned with robustness in the latter sense, which we further sub-categorize based 
on the kind of input perturbations under consideration. We distinguish on the one hand \emph{noise distorsions} that include for example Gaussian noise, 
JPEG compression artifacts but also adversarial examples \cite{szegedy_intriguing_2013} and on the other hand \emph{transformative distorsions} 
comprising image transformations such as rotations and crops. 
A particular challenge arises from the fact that the test data might exhibit distortions 
that have not been encountered by the network during training both in terms of distorsion strength as well as in terms of distorsion types. Therefore we strive to develop methods that ideally lead to \emph{generalizable robustness} beyond distorsions seen during training.

One way of increasing the robustness against input perturbations is to use data augmentation (DA) \cite{krizhevsky2012}. In fact, DA 
by adding perturbed copies of existing data samples is by now an established method and has been shown to greatly increase 
the generalizability of a given model \cite{yaeger_effective_1997,taylor_improving_2017}. DA has two aspects: On the one hand, 
it enlarges the available training data to achieve a better generalization performance and on the other hand it increases the model's robustness 
against transformations used for DA. 
As we will see, this robustness, however, is highly specific to the kind and the particular characteristics of the perturbations used for DA. 
In particular, in the worst case DA can degrade the model performance for unperturbed inputs compared to the baseline model performance.

An alternative approach, which will be explored further in this work, was put forward under the name of stability training (ST) by Zheng et al \cite{zheng_improving_2016}. 
Instead of adding distortion instances to the training corpus, ST feeds the perturbed and the reference sample to the network simultaneously and 
introduces a consistency constraint as additional optimization objective that tries to align the network's outputs of the perturbed image and the reference sample.
By extending the original work beyond Gaussian noise perturbation considered in the original work \cite{zheng_improving_2016}, we propose ST as a general-purpose 
alternative to DA that offers similar advantages with limited negative side effects. We also introduce modifications to the method, which mitigate 
weaknesses and extend its applicability.
In particular, our contributions can be summarized as follows:
\begin{itemize}
    \item We establish stability training as a competitive alternative to data augmentation, which produces comparable or superior robustness improvements 
    across a wide range of distortion types, while exhibiting significantly lower risk to deteriorate performance compared to an unstabilized baseline model.  To this end, we present a detailed analysis of the robustness of 
    both data augmentation and stability training when trained/tested on specific distorsion types.
   
    \item We propose a symmetrical stability objective that increases the method's effectiveness in learning from data transformations, like rotations, that 
    do not distinguish an unperturbed reference sample. The modified objective offers superior performance to data augmentation in scenarios that are 
    likely to be encountered in real-world applications.
    
    \item We evaluate stability training as an alternative to adversarial training to increase robustness against adversarial examples generated via the fast gradient 
    sign method \cite{goodfellow_explaining_2015}.
    
    \item As an outlook we demonstrate the prospects of using multiple distorsion types simultaneously to further improve the robustness properties.
\end{itemize}

\section{Stability Training}

Stability training aims to stabilize predictions of a deep neural network in
response to small input distortions. The idea behind this approach is that an
input \(x'\) that is similar to \(x\) and semantically equivalent ought to
produce similar outputs \(f_\theta(x')\), where $\theta$ denotes the trainable
parameters of the neural network. The full optimization objective is then
defined as a composite loss function of the original task $\Lzero$, for example
cross-entropy between the network's prediction \(y = f_\theta(x)\) and the
(one-hot encoded) ground truth label \(\hat{y}\), and a separate stability
objective $\Lstab$ which enforces the consistency constraint. More explicitly,
given an original image \(x\) and a perturbed version \(x'\) of it, the
combined training objective is then given by

\begin{equation}
        L(x, x',\hat{y}; \theta) = \Lzero(x,\hat{y}; \theta) + \alpha \Lstab(x, x'; \theta)\,, \label{eqn:stabilityobjective}
\end{equation}
where the stability loss is defined via
\begin{equation}
        \Lstab(x, x'; \theta) = D(f_\theta(x), f_\theta(x')), \label{eqn:stabilityobjective-defined}
\end{equation}

and the hyperparameter \(\alpha\) adjusts the relative importance of the two loss components.

The choice of a distance function \(D\) is task-specific. Whereas for
regression tasks the $L_2$-distance is a straightforward choice, in a
classification setting with $C$ classes, where 
\begin{equation}
    \Lzero(x,\hat{y}; \theta)=-\sum_{j=1}^C \hat{y}_j P(y_j|x;\theta)\,,
\end{equation}
the Kullback-Leibler (KL) divergence as considered in
\cite{zheng_improving_2016} represents a canonical way of comparing likelihoods of original 
and distorted inputs
\begin{equation}
D_\text{orig}(y, y') = D_{\text{KL}}(y\!\parallel\!y')\,.
\label{eq:Dnosym}
\end{equation}
The stability loss function is then given by 
\begin{align}
\Lstab(x, x';\theta) &= D_{\text{KL}}(f_\theta(x) \parallel  f_\theta(x'))\nonumber\\
&= -\sum_j P(y_j|x;\theta) \text{ log} \left(\frac{P(y_j|x;\theta)}{P(y_j|x';\theta)}\right)\,.
\end{align}
As the KL-divergence is not symmetric with respect to its arguments, using it as
distance measure is most appropriate in situations, where the reference sample
can be clearly distinguished as undistorted from the modified copy, as it is the case for
distortions we categorize as coming from the \emph{noise} category. For \emph{transformative}
distorsions such as rotations it
turns out to be beneficial to consider a symmetrized stability term, i.e.\

\begin{equation}
\label{eq:Dsym}
    D_\text{sym}(y, y') = \tfrac{1}{2} \left(D_\text{KL}(y\!\parallel\! y')+D_\text{KL}(y'\!\parallel\! y)\right)\,.
\end{equation}

At this point it is worthwhile to point out a crucial difference between
stability training and DA: In the DA-setting the
neural network is trained on perturbed input samples while the loss is the
original loss $\Lzero$ evaluated using the label of the reference image.
On the contrary, stability training decouples solving the
original task from stabilizing class predictions. This is achieved by
evaluating the original loss $\Lzero$ only for the unperturbed samples while the
perturbed samples only enter the consistency-enforcing stability loss term
(\ref{eqn:stabilityobjective-defined}). This construction reduces the potential
negative side-effects of DA, where for certain hyperparameter
choices DA can worsen performance on the original unperturbed
dataset compared to a baseline model. Note that stability training leads to
increased memory requirements as the results for passing the original batch and
the perturbed batch have to be held in memory simultaneously.
\section{Related Work}
This work builds on the original article on ST \cite{zheng_improving_2016} 
that stands out from the literature as one of the few works that explicitly addresses increasing the robustness a neural network against generic input perturbations. We extend their work by considering train distorsions beyond Gaussian noise and a symmetric stability objective to increase the method's performance for transformative distorsions.

There is a large body of works on related methods in the context of semi-supervised learning, see \cite{zhu2005semi,Chapelle:2010:SL:1841234} for reviews. The most relevant works for the present context can be subsumed as consistency/smoothness enforcing methods. The common idea in all cases is to impose a consistency constraint \cite{oliver2018realistic} to enforce similar classification behavior for the original and perturbed input. On the labeled subset of the data both the original loss and the consistency loss can be evaluated, whereas on the unlabeled subset only the consistency loss is imposed. The focus in these works lies on incorporating information from the unlabeled subset to increase the model performance but none of them considered the aspect of robustness with respect to input distorsions. The main difference between the different methods lies in the way how the consistency constraint is implemented \cite{Bachman2014LearningWP,laine2016temporal,sajjadi2016,tarvainen2017mean,Miyato2018VirtualAT,xie2019unsupervised,berthelot2019mixmatch}.

In the domain of DA, Rajput et al \cite{rajput2019does} recently presented a first theoretical investigation of the robustness properties of DA. On the practical side, there have been proposals for more elaborate implementations of DA that try to circumvent the need for dataset-specific hyperparameter searches by appropriate meta search algorithms \cite{cubuk2018autoaugment,zhang2019learning}, which focusing, however, on model performance rather than robustness. In fact, it would represent an interesting line of research to investigate also these techniques from the robustness point of view to see in detail how a far a DA strategy tailored for robustness can get. Mixup \cite{zhang2017mixup} can be seen as an extension to DA in the sense that not only on perturbed input samples but rather on convex combinations of input samples and the corresponding labels are used during training. Recent extensions such as \cite{hataya2019unifying,verma2019interpolation,berthelot2019mixmatch} incorporate also a consistency constraint for stabilization.
\section{Experimental Setup}
\subsection{Tiny ImageNet Full-Sized (TIFS) Dataset} 

We base our experiments on a dataset inspired by the Tiny ImageNet dataset
\cite{tinyimagenet}. The latter represents a reduced version of the original
ImageNet dataset with only 200 instead of the original 1000 ImageNet classes
and 500 samples per class downsized to resolution $64\times 64$. As we wanted
to keep the advantage of the reduced computational demands of Tiny ImageNet
while working on full ImageNet resolutions, we decided to design our custom
TIFS dataset along the lines of Tiny ImageNet but keeping the original samples
from the ILSVRC2012 ImageNet dataset\footnote{A corresponding preprocessing
script is available at \url{https://gist.github.com/nstrodt/bd270131160f02564f0165e888976471}.}. As no labels are provided
for the official ImageNet test set, we used the images from the original validation set
as the basis for the TIFS test set and split the original training set in a cross-validation
fashion, such that for each class we randomly assigned 450 samples to the
training set and 50 to the validation set.

Images contained in the ImageNet dataset and thus in the TIFS dataset come at
varying sizes. We preprocess the dataset such that every image in the dataset
is first resized, such that the shortest side is 256 pixels in length. Next, we
crop out a 224 pixels wide quadratic area, such that the center points of the
crop and the image coincide. The image is then converted from RGB integer color
values ranging 0 to 255, to floating-point values ranging 0 to 1. Distortions
are applied at this point. Finally, we normalize and whiten the (undistorted) 
images such that the channel-wise mean and variance across the entire dataset are
0 and 1, respectively.

\subsection{Model Architecture and Optimization}

In our experiments, we use a deep convolutional residual network (ResNet18)
\cite{he_deep_2016} as prototype for a state-of-the-art convolutional neural
network that is presently used predominantly in computer vision applications.
To optimize our model, we use mini-batch stochastic gradient descent with
Nesterov accelerated momentum \cite{nesterov_method_1983} and batch
normalization \cite{ioffe_batch_2015}. We use the default \emph{torchvision}
implementation of ResNet18 as supplied with Pytorch \cite{paszke2017automatic}
and use randomly initialized weights rather than pretrained weights that
already include DA during pretraining.

We find that introducing the stability objective late during the training phase
produces similar results to applying it from the beginning and therefore
decided to use the following experimental procedure: We initially train a model
for 30 epochs on the original training set with no distortions added to the
preprocessing procedure. We save the model at each epoch and select the model
with the highest performance on the undistorted validation set. This model
serves as baseline model in the following experiments. All subsequent models
are initialized with the weights of the baseline model. In an individual run using either ST or DA, the model is trained in the same fashion as the
baseline model, except for a limited number of 10 epochs. For a selected distorsion
we train a number of models according by varying distorsion hyperparameters. 
Similarly to the training of the baseline
model we use the model performance on the undistorted validation set to perform
early stopping, i.e.\ for a specific hyperparameter choice we select the model
with the highest score on the undistorted validation set. We keep a fixed learning
rate of \(0.01\) and a batch size of 128 for all experiments.

\begin{table}[ht]
  \caption{Considered distortion types. ``Type'' denotes the assignment either to N(oise) or to T(ransformative) distorsions. The column
   ``Practical'' reflect a subjectively selected parameter value likely to be encountered for distorsions in real world applications.}
  \label{tab:distortioncategories}
  \centering
  \begin{tabular}{|l|l|l|}
    \hline
    Distortion (Type)& Parameter& Practical\\
    \hline
    \hline
    Gaussian noise (N)& standard deviation $\sigma$& 0.05\\
    JPEG compression (N)& quality $q$& 30\\
    Thumbnail resizing (N) &pixel size $A$ & 150\\
    FGSM Adversarial examples (N) & strength $\epsilon$ & 0.001\\\hline
    Rotation (T)&max.\ rotation $\rho$&30\\
    Random cropping (T)&pixel offset $C$& 3\\
    \hline
  \end{tabular}
\end{table}

We consider a range of different distortion types that are summarized in
Tab.~\ref{tab:distortioncategories}. As already discussed above, we broadly
categorize distortions as undirected noise distorsions
or as transformative distortions arising from geometric transformations
of the input image. For later reference we subjectively identified a parameter
value for \emph{practical} distorsions, that reflects a distorsion strength
likely to be encountered in real world applications or in certain settings,
like FGSM, a distorsion strength where the distortion is on the verge of being
detectable by a human, but has not yet surpassed that threshold.

We consider the following distorsions: Gaussian noise adds pixel-wise
independent Gaussian noise to the image such
that \(x'_k = x_k + \epsilon_k,\;\epsilon_k \sim \mathcal{N}(0, \sigma^2)\),
where $k$ indicates the index of the pixel in $x$ and the standard deviation
$\sigma$ serves as hyperparameter. JPEG compression is an image
compression algorithm that aims to minimize file size while maximizing retained
semantic meaning. It offers a quality level $q \in [1, 100]$ indicating how much
image quality is favored over file size. Thumbnail resizing crops the image
down to a quadratic area with side-length $A$ and afterwards resizes the image
to its original dimensions. This introduces interpolation artifacts that
increase in severity with lower values of $A$. We choose FGSM
\cite{goodfellow_explaining_2015} as an example method to produce adversarial
examples for its simplicity. An example is generated via \(x' = x + \epsilon
\text{ sign}\left(\nabla_x L(x, \hat{y}; \theta)\right)\), where $\epsilon$ is
a strength parameter and $L$ is the full loss function. The rotation distortion
represents a random rotation up to $\rho$ degrees. Random cropping, unlike the
other distortions considered here, changes the common preprocessing procedure.
Usually the center point of the final crop and of the image coincide. This
distortion offers a parameter $C$ such that these points are displaced by up to
$C$ pixels in all four directions.

For DA we introduce the probability $p$ that the selected augmentation will be
applied as an additional hyperparameter. This is the default setup in which
DA is applied in practical applications and allows to mitigate
effects of catastrophic forgetting \cite{french_catastrophic_1999}, i.e.\ a
performance deterioration on the original undistorted dataset that is observed
for certain hyperparameter choices. The corresponding
hyperparameter for ST is the coefficient $\alpha$, see Eq.~\ref{eqn:stabilityobjective}, that sets the relative scale
of the consistency loss term compared to the original loss.

We conclude this section with a remark on the intricacies of comparing ST and DA. 
If one uses the same batch size
in both cases and fixes the number of epochs to the same value in both
settings, one the one hand ST is fed twice the amount of raw data. On the other
hand, if DA is trained using the same number of examples i.e.\ by doubling the number of
epochs compared to ST, it allows DA to make the double number of label uses compared to ST.
Whereas the ST performance typically stabilizes rather quickly during training,
increasing the number of training epochs in the DA setting has implications on
the robustness properties of the model that will be discussed qualitatively
below.

\section{Results}

\begin{table}[t]
  \centering
  \caption[Hyperparameter ranges]{Hyperparameter ranges used during training.}
  \begin{tabular}{|l|c|c|}
    \hline
    Hyperparameter     & [start,end,\# points] &  Scale\\
    \hline\hline
    ST: relative weight \(\alpha\) & [0.01, 10.0, 3] & logarithmic  \\
    DA: transformation probability $p$ & [0.5, 1.0, 2] & linear  \\
    \hline
    Gaussian noise \(\sigma\) & [0.01, 1.0, 4] & logarithmic \\
    JPEG compression \(q\) & [90, 10, 3] & linear \\
    thumbnail resizing \(A\) & [20, 200, 3] & linear \\
    FGSM Adversarial examples \(\epsilon\) & [0.001, 1.0, 7] & logarithmic \\
    rotation \(\rho\) & [0, 180, 3] & linear \\
    random cropping \(C\) & [0, 15, 3] & linear \\
    \hline
  \end{tabular}
  \label{tab:hyperparamgridsearch}
\end{table}

In the following, we refer to the type of distortion to be used for
regularization in conjunction with ST or DA during training(testing) as
\emph{training (test) distortion}. In our experiments we train models with a variety of
training distortions from both distorsion types and evaluate their performance
on various test distortions, see 
Tab.~\ref{tab:hyperparamgridsearch} for the hyperparameter ranges considered during training. The
validation set is only used for early stopping based on the model performance
on the undistorted validation set as described above. With the scenario of
unknown test perturbation in mind, we present the corresponding test set
performances for different hyperparameter choices. To guide the eye we
typically report both for ST and for DA the
performance of the model that performed best/worst at the specified
practical distorsion level but we do not perform any form of model selection with respect 
to the distorsion hyperparameters. Finally, for ST we also
report the performance range upper and lower bounds of any training setting at
any test setting as a shaded area. This is used as an unbiased measure for the
robustness of ST with respect to hyperparameter choices. In our
results, we distinguish between those experiments where training and test
distortion are of the same type and those where they differ. On the one hand,
this allows us to investigate how well a method can increase robustness against
a given distortion type and, on the other hand, it shows what side effects on other
distortions this might induce.

\begin{figure*}[ht]
     \centering
     \begin{subfigure}[b]{0.46\textwidth}
         \centering
         \includegraphics[width=\textwidth]{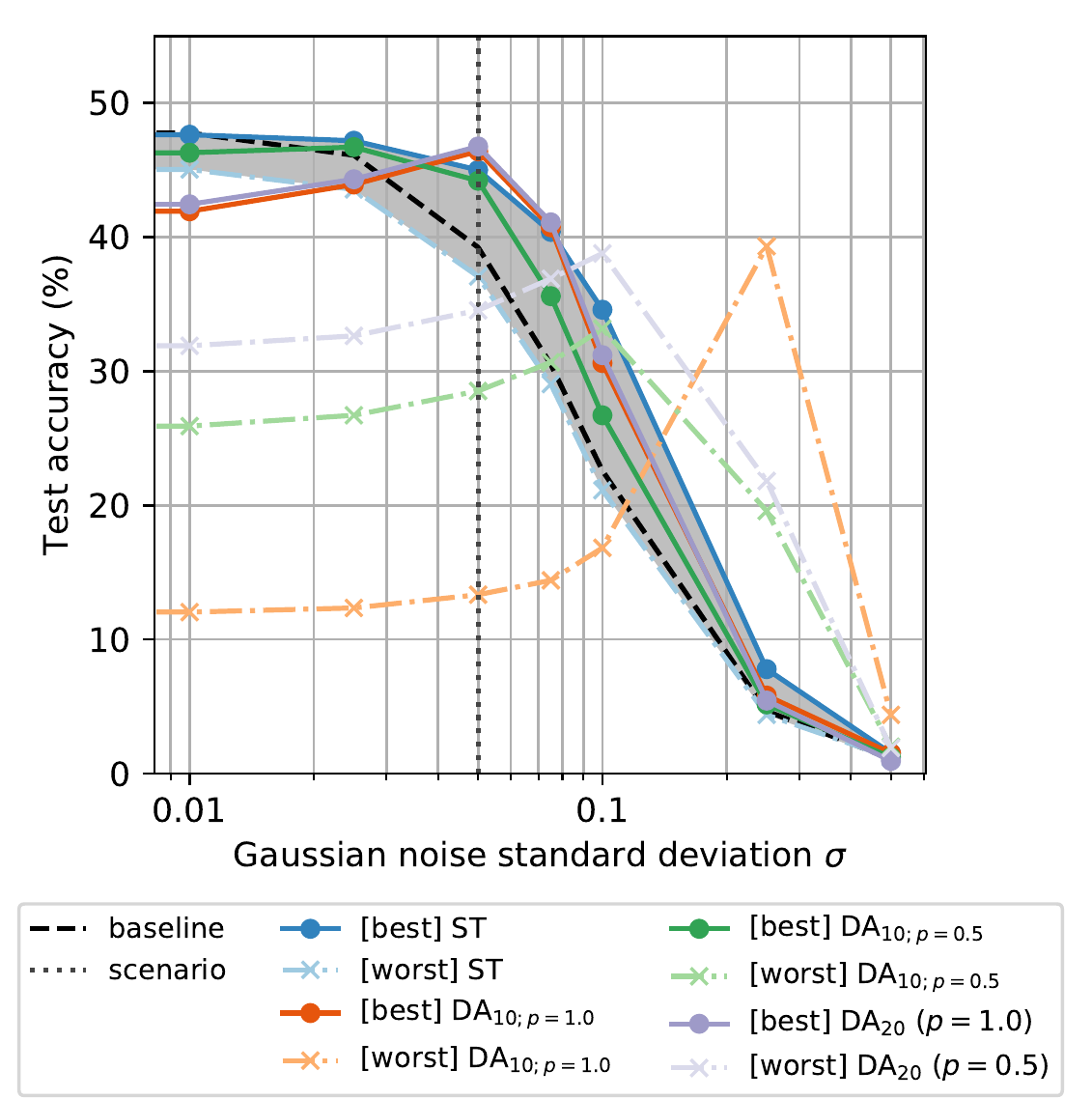}
         \caption{Gauss $\to$ Gauss}
         \label{fig:gauss-gauss}
     \end{subfigure}
     \hfill
     \begin{subfigure}[b]{0.46\textwidth}
         \centering
         \includegraphics[width=\textwidth]{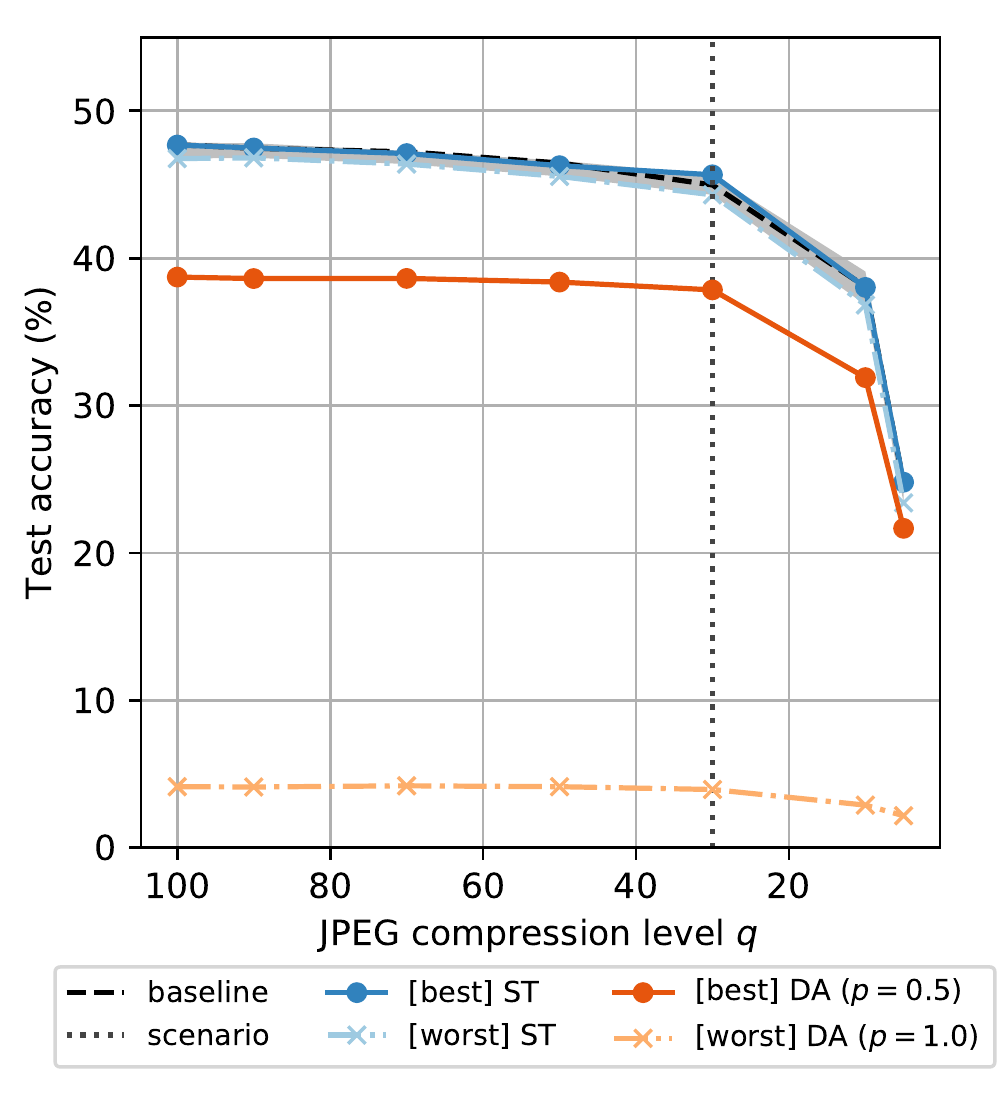}
         \caption{Thumbnail $\to$ JPEG}
         \label{fig:thumb-jpeg}
     \end{subfigure}        
        \caption{Robustness properties of stability training (ST) vs.\ data
        augmentation (DA) for noise distorsions.  The figure captions such as ``Thumbnail $\to$ JPEG'' 
        refer to thumbnail resizing as training and JPEG as test distorsion. We typically report the
        performance of the best/worse model selected at the practical distorsion
        level. Subscripts refer to subsets of hyperparameters, i.e.\ [best]
        $\text{DA}_{10;p=0.5}$ refers to the best model among the set of all
        models with hyperparameter $p=0.5$ trained for 10 epochs. On the
        contrary, values in parentheses denote hyperparameters of the selected
        best/worst model.}
        
        \label{fig:noise-cat}
\end{figure*}

\begin{figure*}[ht]
  \centering
  \begin{subfigure}[b]{0.46\textwidth}
    \centering
    \includegraphics[width=\textwidth]{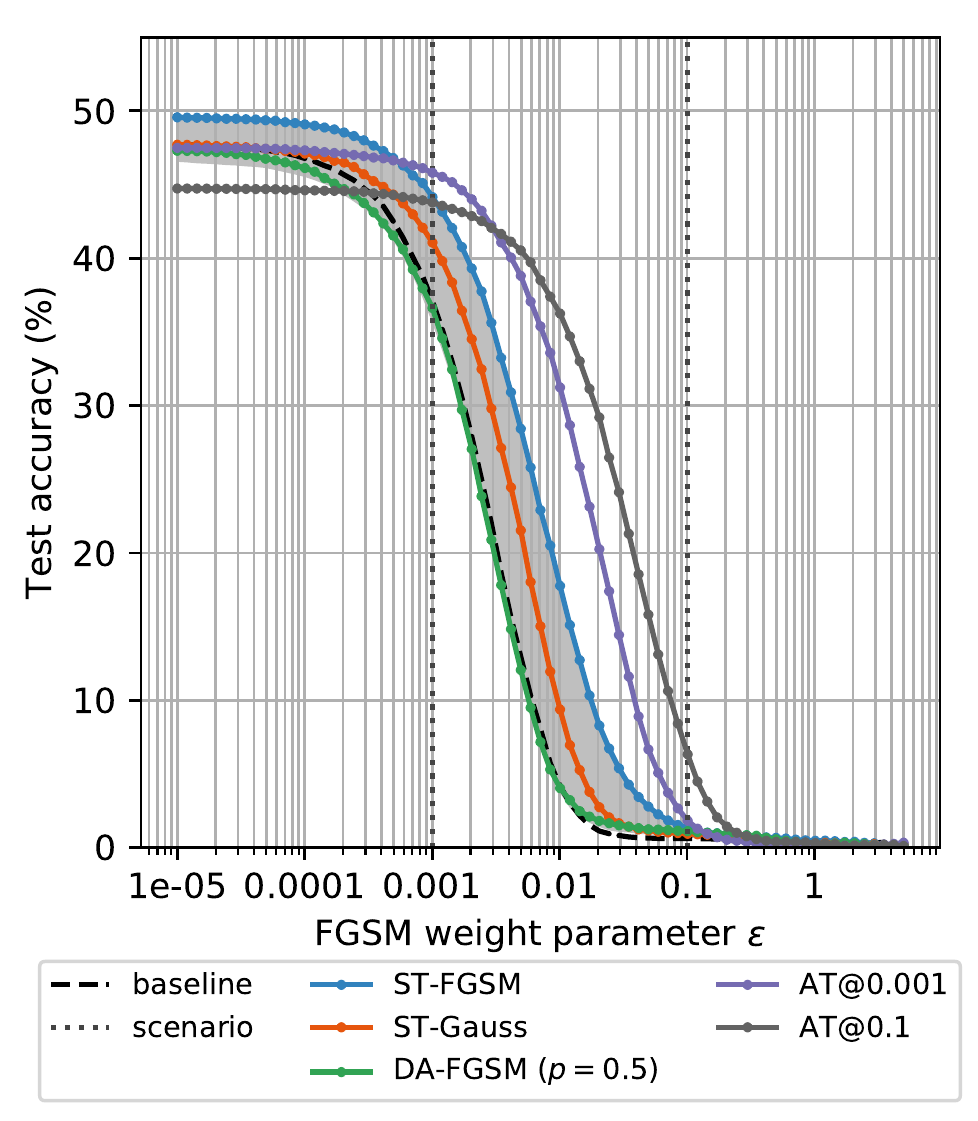}
    \caption{FGSM $\to$ FGSM }
    \label{fig:fgsm-fgsm}
\end{subfigure}
  \hfill
  \begin{subfigure}[b]{0.46\textwidth}
    \centering
    \includegraphics[width=\textwidth]{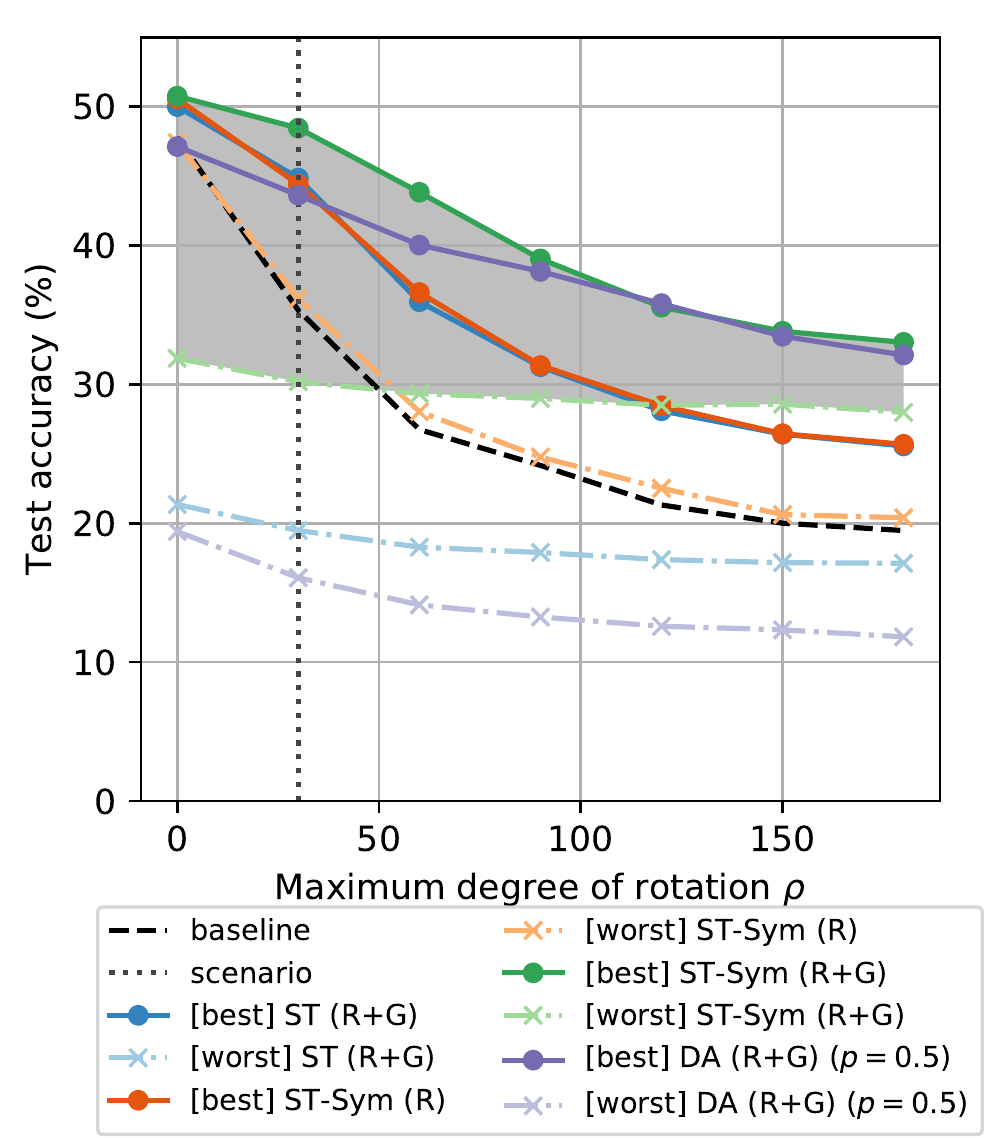}
    \caption{Gauss + Rotation $\to$ Rotation}
    \label{fig:gauss_rot-rot}
  \end{subfigure}  
     \caption{Robustness properties of stability training (ST) vs. data augmentation
    (DA) for FGSM adversarial examples and using multiple training distorsions. We use the same color-coding and nomenclature as in Fig.~\ref{fig:noise-cat}.}
     \label{fig:noise-cat2}
\end{figure*}

In Figs.~\ref{fig:gauss-gauss}, \ref{fig:thumb-jpeg}, and \ref{fig:fgsm-fgsm} we compare the robustness properties of ST and DA for three different setups within the category
of noise distorsions. Fig.~\ref{fig:gauss-gauss} shows the robustness using
Gaussian noise as training distorsion, which corresponds to the setting considered
in the original ST paper\cite{zheng_improving_2016}. We observe
in Fig.~\ref{fig:gauss-gauss} that DA without reintroducing the
reference image ($p = 1.0$) outperforms ST only with its peak
performance at the chosen practical distortion level but shows a worse performance compared to ST elsewhere. Even more importantly, an unfavorable
hyperparameter choice for DA can lead to catastrophic
forgetting, which can be seen quite drastically for the worst-performing DA
models. This issue can be mitigated to some degree by the reintroduction of the
reference image during training, here for $p=0.5$, which, however results in
worse performance (compared to DA with $p=1.0$) at the practical distortion
level. We also use this setting to illustrate the impact of an increased number
of training epochs. Here we show results for DA trained for 20 epochs,
which corresponds to the same number of batches seen during ST
but double the number of labeled examples. Interestingly, increasing the number
of training epochs does not result in a model with better performance or
robustness. It only marginally improves the best-performing result found at 10
epochs. Even though the worst-performing model is improved by a sizable amount
from 10 to 20 epochs, the ST range as illustrated by the area shaded in gray in
Fig.~\ref{fig:gauss-gauss} is still considerably smaller than the range of results
obtained from models trained with DA. This leaves ST as the most favorable
choice for stabilizing against Gaussian noise consistent with the claims in
the original paper\cite{zheng_improving_2016}. The qualitative findings
concerning the impact of an increased number of epochs and the comparison of
$p=0.5$ and $p=1.0$ represent a general pattern observed during all of our
experiments. In the following plots we therefore conventionally show only the
worst/best DA performance treating $p$ as an additional hyperparameter and
restrict ourselves to the setting of 10 training epochs for DA.

Fig.~\ref{fig:thumb-jpeg} illustrates the cross-distorsion performance of both
DA and ST. In this particular example, we trained with thumbnail rescaling and
evaluated performance on JPEG compression. This setting resembles real-world
applications where the model might encounter distorsion types unseen during
training. As training and test distortion do not have many characteristics in
common, we do not expect to see noticeable performance improvements via either
method. Importantly, however, we can observe, that DA can
severely harm the model's robustness against distortion types that were not
present during training. Even the best possible model trained via data
augmentation performed with a deficit of nearly 10\% compared to baseline.
ST, on the other hand, did not show any worsening impact. This
example should not convey the impression that ST cannot acquire robustness from cross-category
distorsions. In fact, for any test distorsion in the noise category, ST performs best 
using Gaussian noise as training distorsion, while DA always favors coinciding training and test distorsions.

We also investigate the prospects of using ST to increase
adversarial robustness. To this end, we dynamically generate adversarial
examples via FGSM and feed them as the perturbed image via either ST, DA or adversarial training (AT) with $\mu = 0.5$
\cite{goodfellow_explaining_2015}, where the latter corresponds to the
optimization objective

\begin{equation}
    L_\text{adv}(x,x',\hat{y}) = \mu \Lzero(x,\hat{y}; \theta) + (1 - \mu) \Lzero(x',\hat{y}; \theta)\,, \label{eqn:adversarial_training}
\end{equation}

where $\mu\in[0.1]$, $x$ denotes the reference sample and $x'$ the adversarial
example and $\hat{y}$ is the ground truth label. In Fig.~\ref{fig:fgsm-fgsm},
we observe that DA is unable to generalize at all, performing
similar to baseline. ST offers a significant improvement
compared to baseline performance across the entire intensity spectrum. Standard
adversarial training offers no to marginal improvements at low intensity
levels, but excels in mid-ranges, where it outperforms ST. To
investigate this further, we also plot the results if we had selected the
best-performing model for an extreme distorsion scenario ($\epsilon = 0.1$)
instead of the regular practical scenario ($\epsilon = 0.001$). Here we see,
that adversarial training sacrifices performance in the low-intensity domain
while gaining in the mid- to high-intensity domains. The ST
performance remains virtually unchanged comparing the best-performing models at extreme 
distorsion to that a practical distorsion level. As the gray ST band indicates, ST performance never drops below baseline performance. Interestingly, even ST with
Gaussian noise leads to an increased adversarial robustness compared to the
baseline performance. We are very well aware of the limitations of robustness
against FGSM as indicator for general adversarial robustness
\cite{DBLP:journals/corr/abs-1902-06705} and merely see our findings as an
indicator the general robustness properties of ST and a
potential direction of future research.

\begin{figure*}[ht]
     \centering
     \begin{subfigure}[b]{0.46\textwidth}
         \centering
         \includegraphics[width=\textwidth]{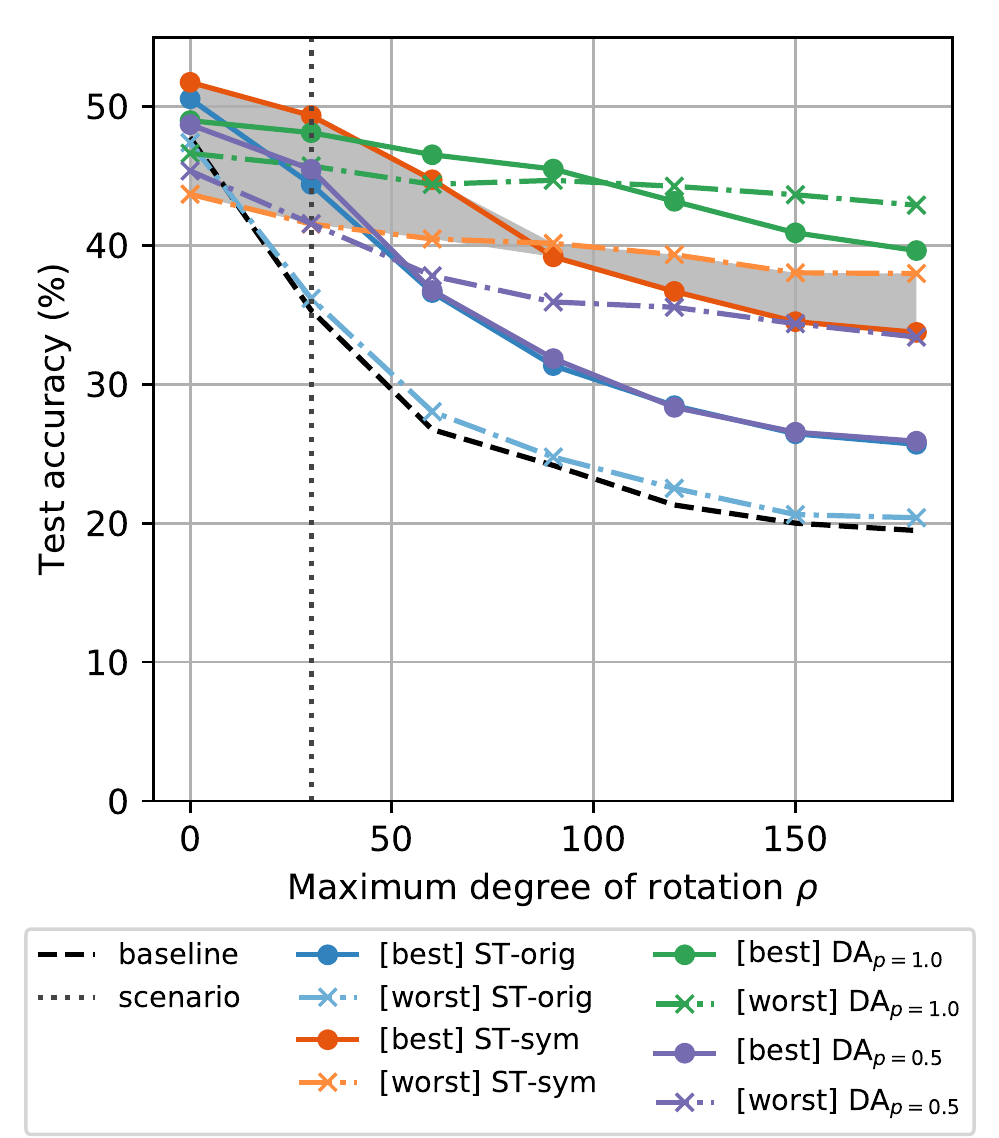}
         \caption{Rotation $\to$ Rotation}
         \label{fig:rot-rot}
     \end{subfigure}
     \hfill
     \begin{subfigure}[b]{0.46\textwidth}
         \centering
         \includegraphics[width=\textwidth]{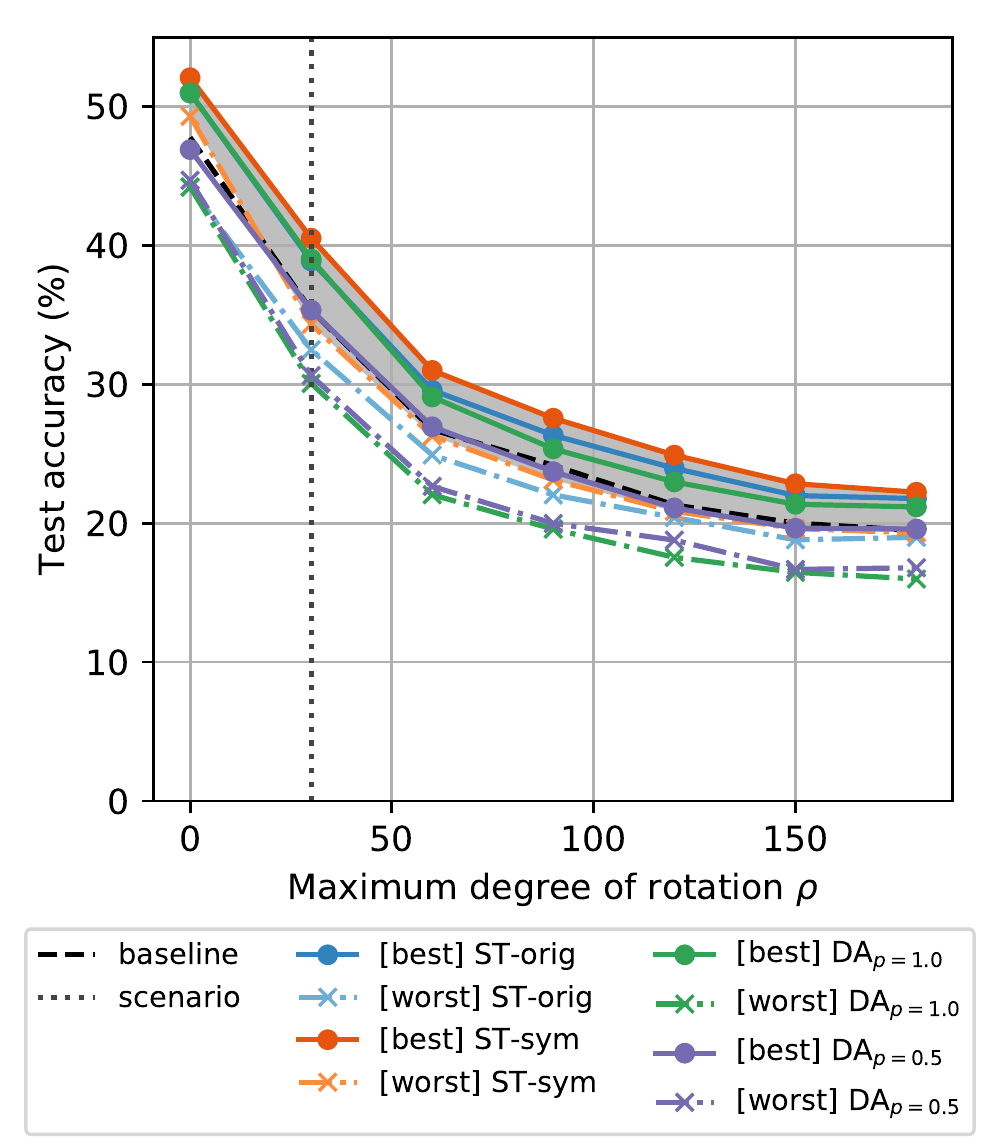}
         \caption{Crops $\to$ Rotation}
         \label{fig:crops-rot}
     \end{subfigure}  
    \caption{Robustness properties of stability training (ST) vs.\ data augmentation (DA) for transformative distorsions. We use the same color-coding and nomenclature as in Fig.~\ref{fig:noise-cat}.}
     \label{fig:trans-cat}
\end{figure*}

Now we turn to the results on transformative distorsions as presented in
Fig.~\ref{fig:trans-cat}. In particular, we compare the performance of the
model trained using the symmetric stability objective from Eq.~\ref{eq:Dsym} to
the performance of those trained with the standard stability objective. In
fact, we observe in Fig.~\ref{fig:rot-rot} that the symmetrized stability
objective does remedy the shortcoming of the original stability objective when
training with rotation. Across all intensity levels $\rho>0$, the symmetrical
objective can significantly increase the model's robustness towards rotation
compared to regular ST. Even the increased performance of the
original ST objective at $\rho=0$ is consistent with expectations as the original ST
loss distinguishes the reference image. We also show
performance for DA ($p=0.5$) with and without ($p=1.0$)
reintroducing the reference image. While the reintroduction of the
reference image showed improvements in Fig.~\ref{fig:gauss-gauss}, we observe the opposite effect in Fig.~\ref{fig:rot-rot}. It is important to
note, that the tested rotation range beyond the practical scenario is
unrealistically large for real-world applications. In the range up to rotations
in the practical scenario, the symmetrical stability loss offers the best
performance of all configurations. However, it is interesting to note that the
performance of the worst symmetrical ST model unlike that of the worst original
ST model drops below the baseline performance in the undistorted case.
Similarly to the noise category experiments, we evaluate the performance across
distortion instances of the different type. This is shown exemplarily in
Fig.~\ref{fig:crops-rot}, where we trained models with varying crops of the
input image and evaluated the resulting performance on rotated images. Again,
we observe that symmetrical ST offers the best performance of
all configurations across the entire intensity spectrum. Also, the gray area
shows that no ST model drops below baseline
performance, while DA does for unfavorable hyperparameter
choices. 

At this point we summarize the results of our single-distorsion experiments: 
Even though we decided to present the results in the form of examples to
illustrate the performance characteristics through the full range of distorsion
strengths, we want to stress that these examples just examplify the more
general picture underlying our investigations, see also the figures in the supplementary material. 
For all considered distorsions
in the setting of identical train and test distorsions ST outperformed DA in
the practical distorsion range in terms of best-model performance with a
considerably smaller hyperparameter dependence. In particular, DA tends to optimize 
robustness by increasing the peak performance at the particular distorsion characteristics
used during training whereas ST typically achieves a stable performance throughout the
whole practical distorsion range, see e.g.\ Fig.~\ref{fig:gauss-gauss}.
ST also shows general
robustness i.e.\ generalizes to unseen distorsions within the same
noise/transformative distorsion category. We also investigated the
use of cross-category training, i.e.\ training using a distorsion from the
noise category while evaluating on the transformative category, but observed no
 noticeable improvements compared to baseline performance.

The more favorable setting and an interesting direction for future research
turned out to be the combination of distorsions from both categories. As ST performs
best for test distorsions of the noise categoy using Gaussian noise during training, we investigate the impact of using both Gaussian noise and rotation during training and compare its performance against models trained solely with the same distortion as used for testing (rotation). As shown in
Fig.~\ref{fig:gauss_rot-rot}, regular ST offers some
improvements across the entire intensity spectrum and shows no difference when
adding Gaussian noise in addition to rotation during training. Data
augmentation performs well when trained solely with rotation across the entire
spectrum and can improve on its performance on mid to high intensity levels by
adding Gaussian noise as an additional regularization. This comes,
however, at the cost of a reduced performance in the practically relevant
intensity range. The symmetrical stability loss offers the best performance
across the entire spectrum of intensities, showcasing again how stability
training is capable to utilize Gaussian noise a universal distortion
approximator to generally improve robustness. Also in this setting, the ST band is still
considerably smaller than the performance range for DA applied using the two
distorsions simultaneously.

\section{Conclusion}
In this work we investigated methods to increase the robustness of deep neural
networks against various kinds of input distortions. To this end we thoroughly
analyzed the prospects of using stability training
\cite{zheng_improving_2016} for this purpose, extending it beyond its original
working domain of stabilization on Gaussian noise to arbitrary input
distortions. We evaluated the proposed method on an ImageNet-scale image
classification task and compared the robustness of models stabilized by
stability training to those stabilized by data augmentation as predominantly
used approach in practical applications to increase both generalization performance
and robustness.

In our experiments we demonstrated that stability training performs on a par
or even outperforms data augmentation in all investigated distortion settings
ranging from noise distortions, including FGSM adversarial examples, to image
transformations. Most importantly, stability training is considerably less
sensitive to hyperparameter choices than data augmentation, whose performance
on the original undistorted dataset may even deteriorate significantly compared to
the baseline model for unfavorable hyperparameter values. Its generalizable robustness 
property makes stability
training a particularly good choice for applications where the specific
characteristics of the distortion to be encountered is unknown and general
robustness is desired. The exploration of using multiple distortion types
jointly for stabilization showed first promising results and represents an
interesting direction for future research.

\section*{Acknowledgements}

This work was supported by the Bundesministerium f\"ur Bildung und Forschung
(BMBF) through the Berlin Big Data Center under Grant 01IS14013A and the Berlin
Center for Machine Learning under Grant 01IS18037I.

\bibliographystyle{splncs04}
\bibliography{stability_training}

\appendix
\section{Supplementary Material}
In this section we present a more complete overview of results for different training and test distorsion that could not be included in the main text due to space constraints.
\newpage
\renewcommand{\figurename}{Supplementary Figure}
\setcounter{figure}{0} 

\begin{figure}[htb]
    \centering
\begin{subfigure}{0.4\textwidth}
  \includegraphics[width=\linewidth]{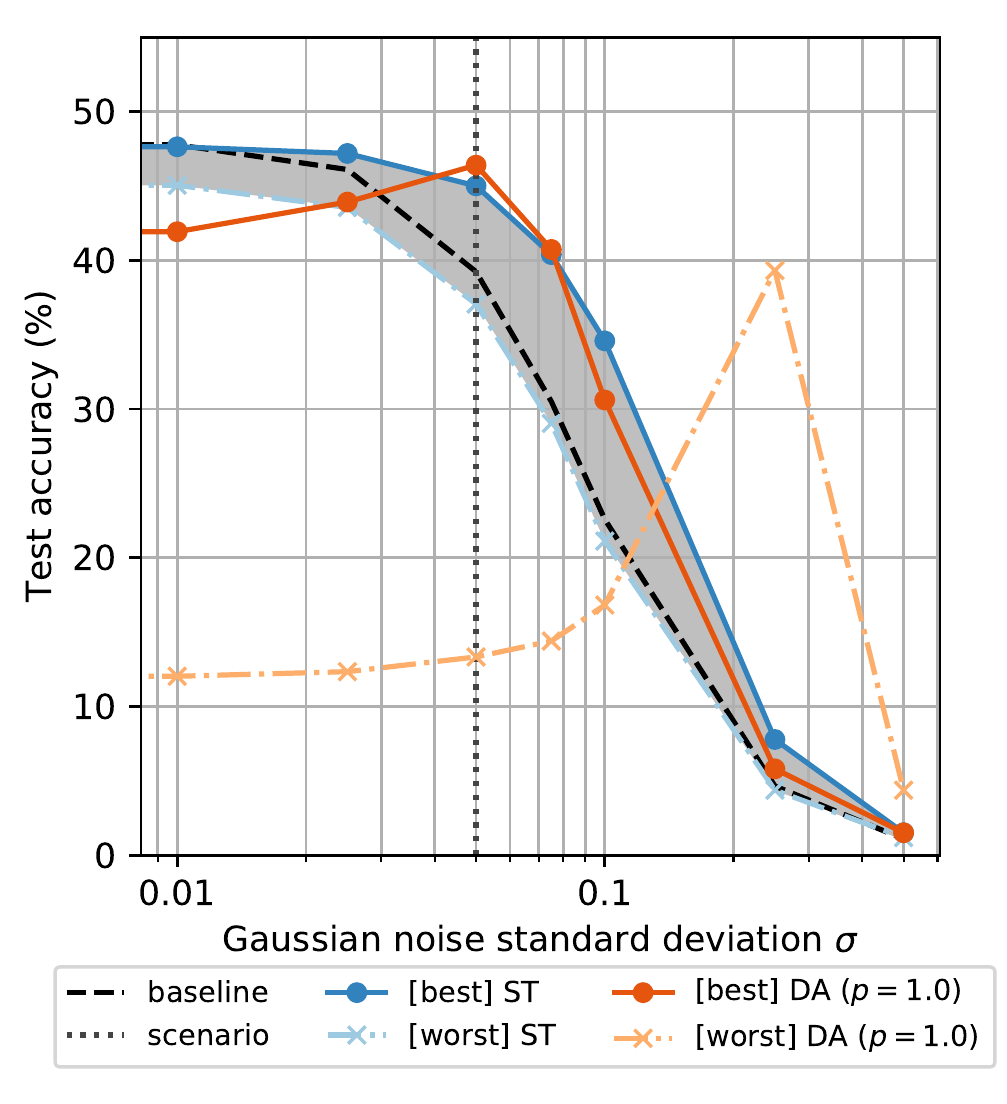}
  \caption{Gauss $\to$ Gauss}
\end{subfigure}\hfil
\begin{subfigure}{0.4\textwidth}
  \includegraphics[width=\linewidth]{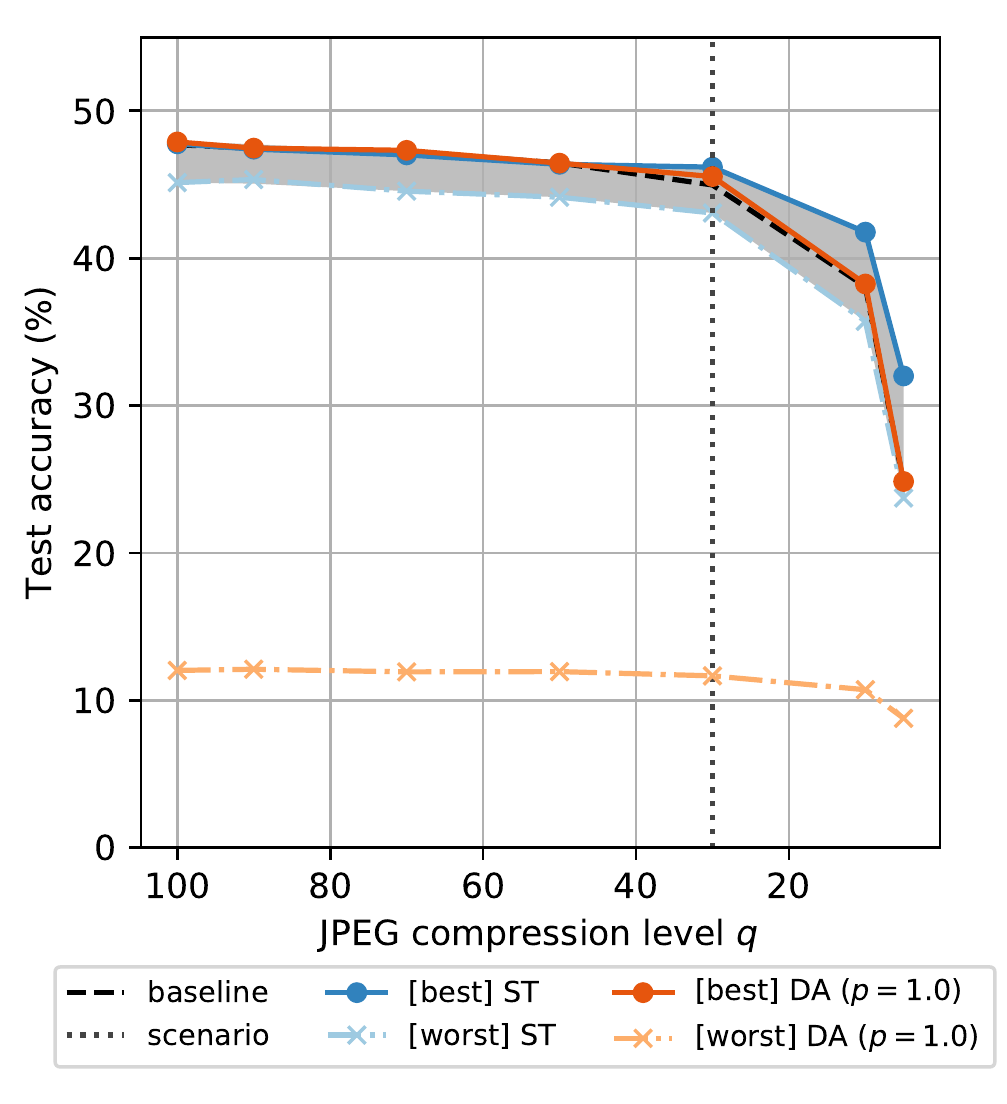}
  \caption{Gauss $\to$ JPEG}
\end{subfigure}
\begin{subfigure}{0.4\textwidth}
	\includegraphics[width=\linewidth]{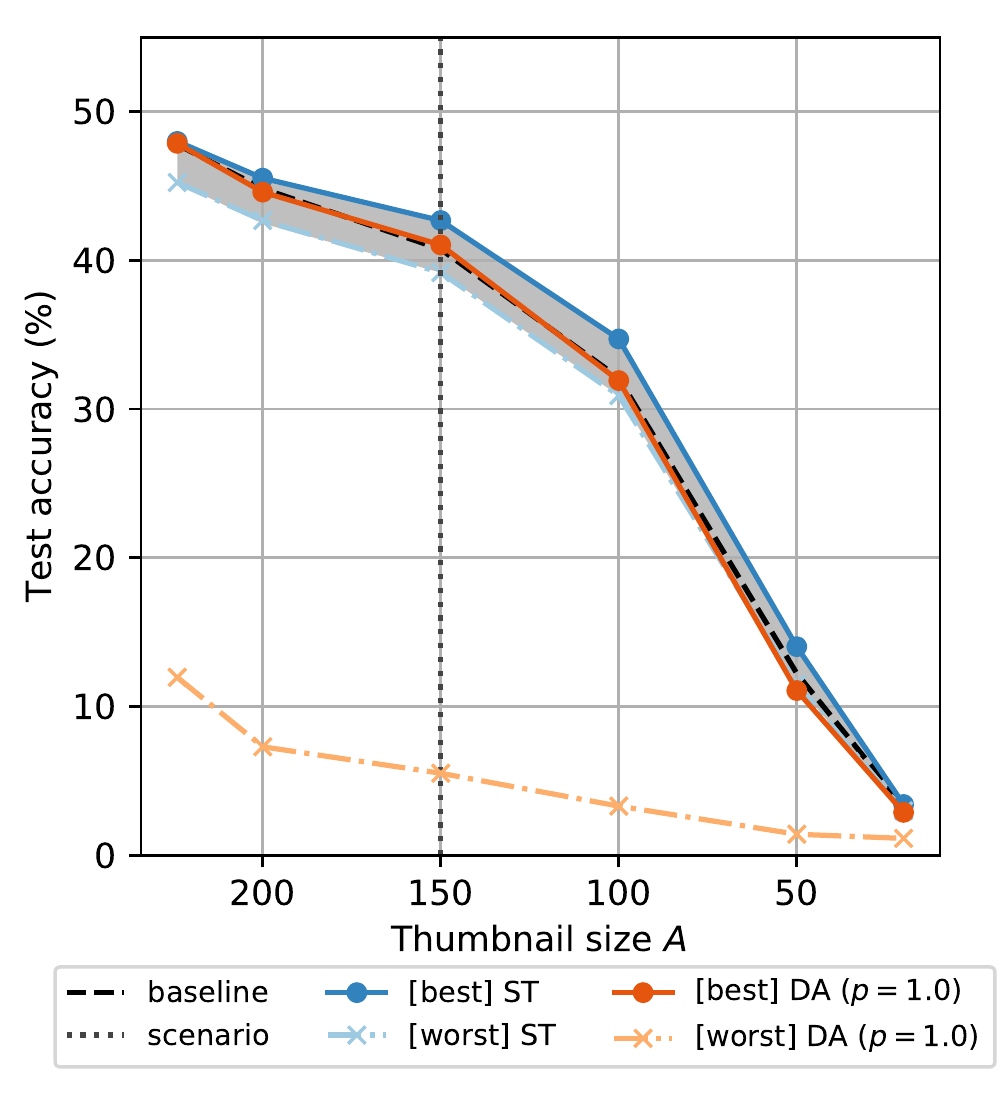}
	\caption{Gauss $\to$ Thumb}
  \end{subfigure}\hfil
  \begin{subfigure}{0.4\textwidth}
	\includegraphics[width=\linewidth]{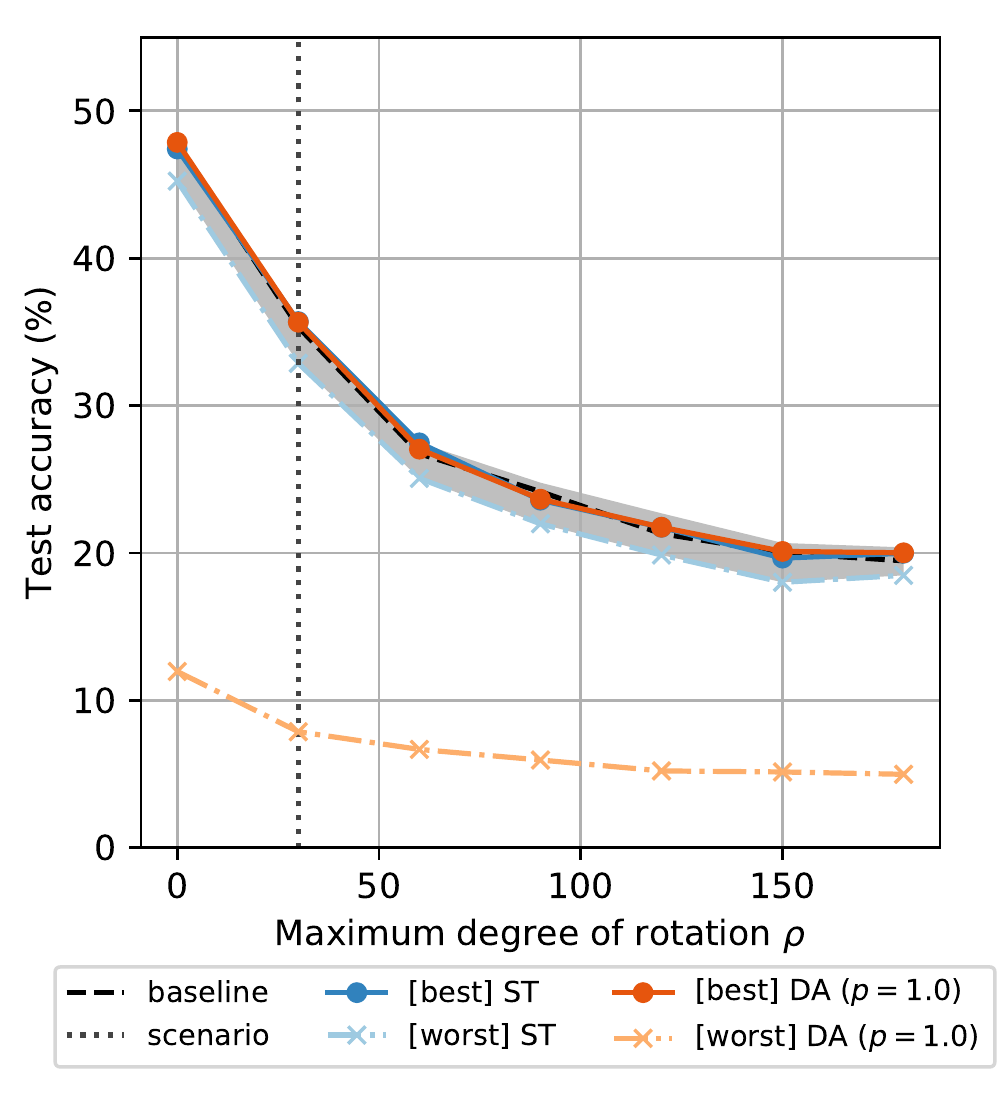}
	\caption{Gauss $\to$ Rotation}
  \end{subfigure}
  \begin{subfigure}{0.4\textwidth}
	\includegraphics[width=\linewidth]{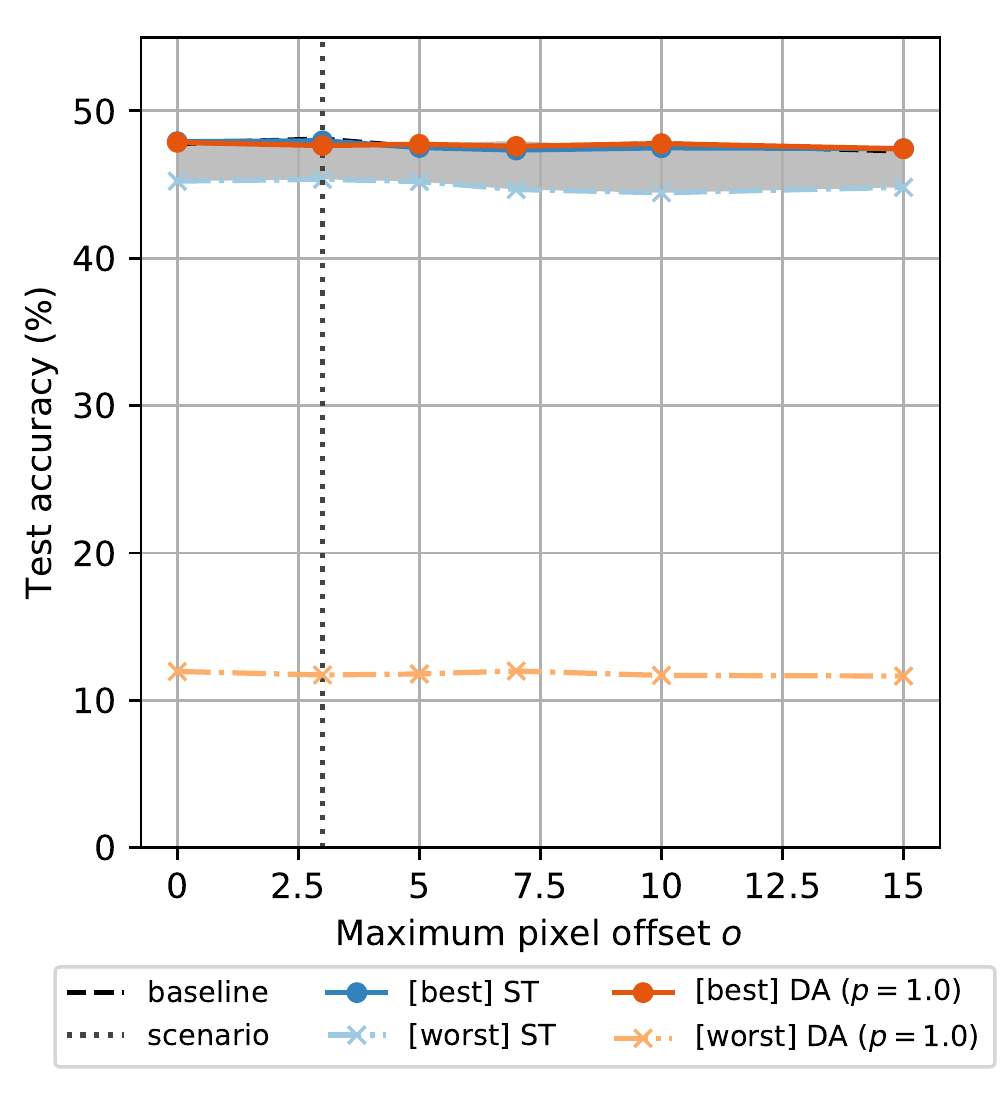}
	\caption{Gauss $\to$ Crops}
  \end{subfigure}\hfil
\caption{Gaussian noise as training distorsion.}
\end{figure}

\begin{figure}[htb]
    \centering
\begin{subfigure}{0.4\textwidth}
  \includegraphics[width=\linewidth]{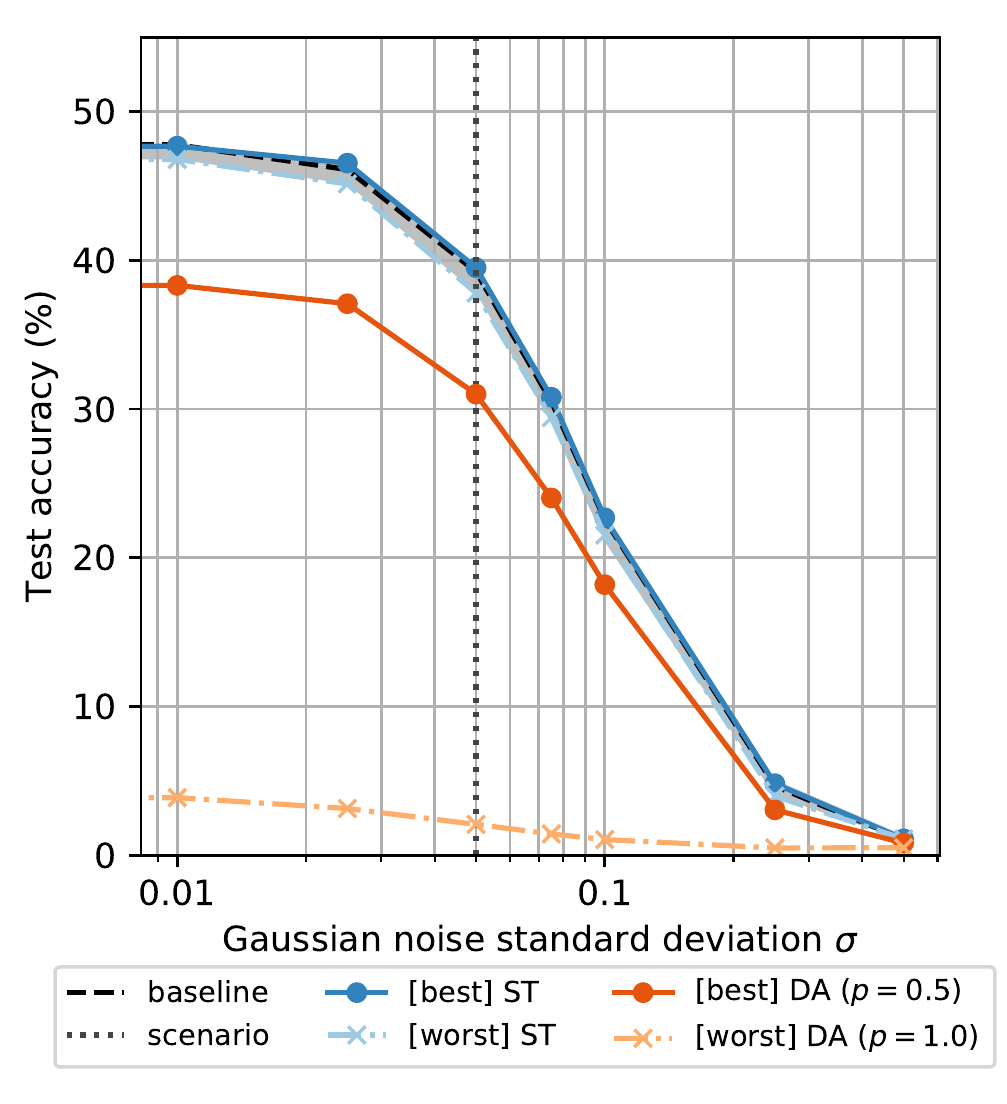}
  \caption{Thumb $\to$ Gauss}
\end{subfigure}\hfil
\begin{subfigure}{0.4\textwidth}
  \includegraphics[width=\linewidth]{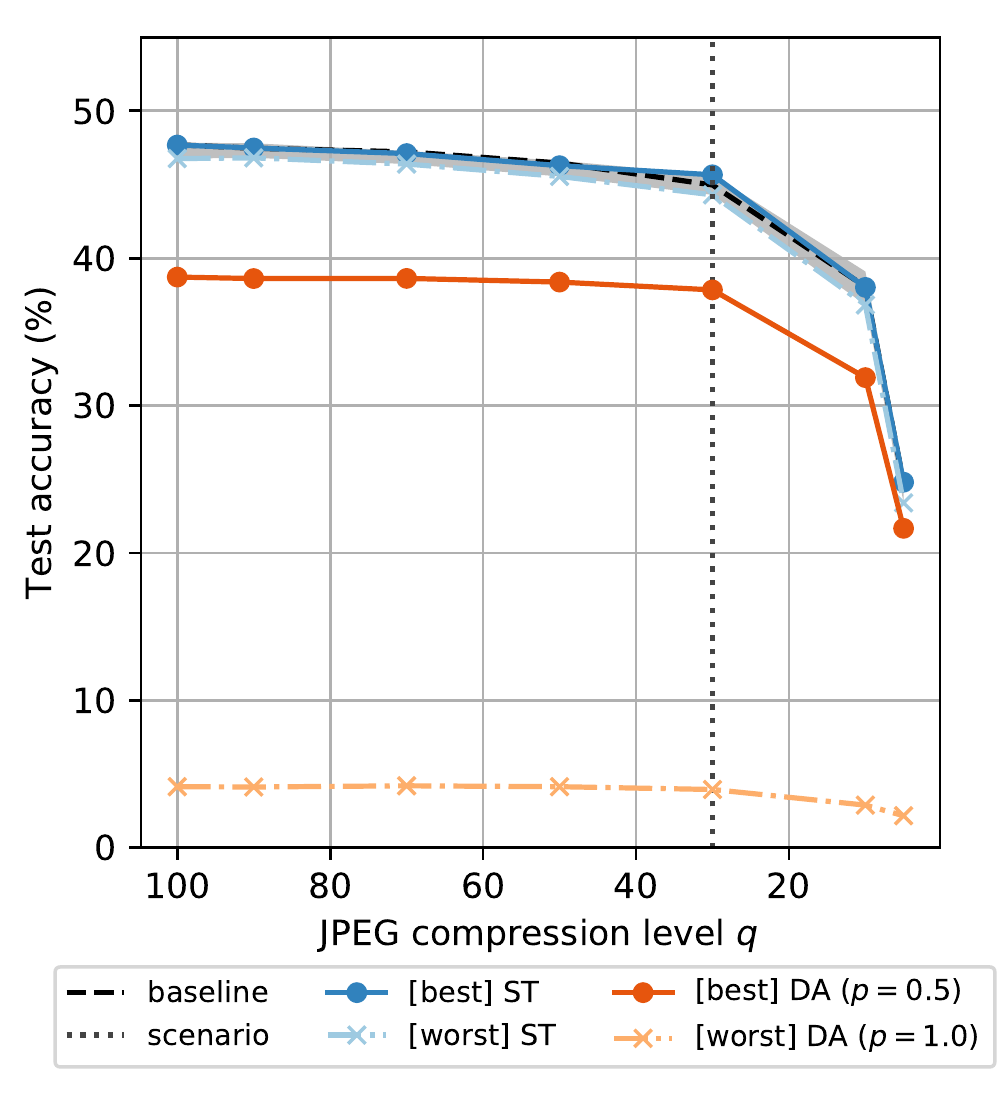}
  \caption{Thumb $\to$ JPEG}
\end{subfigure} 
\begin{subfigure}{0.4\textwidth}
	\includegraphics[width=\linewidth]{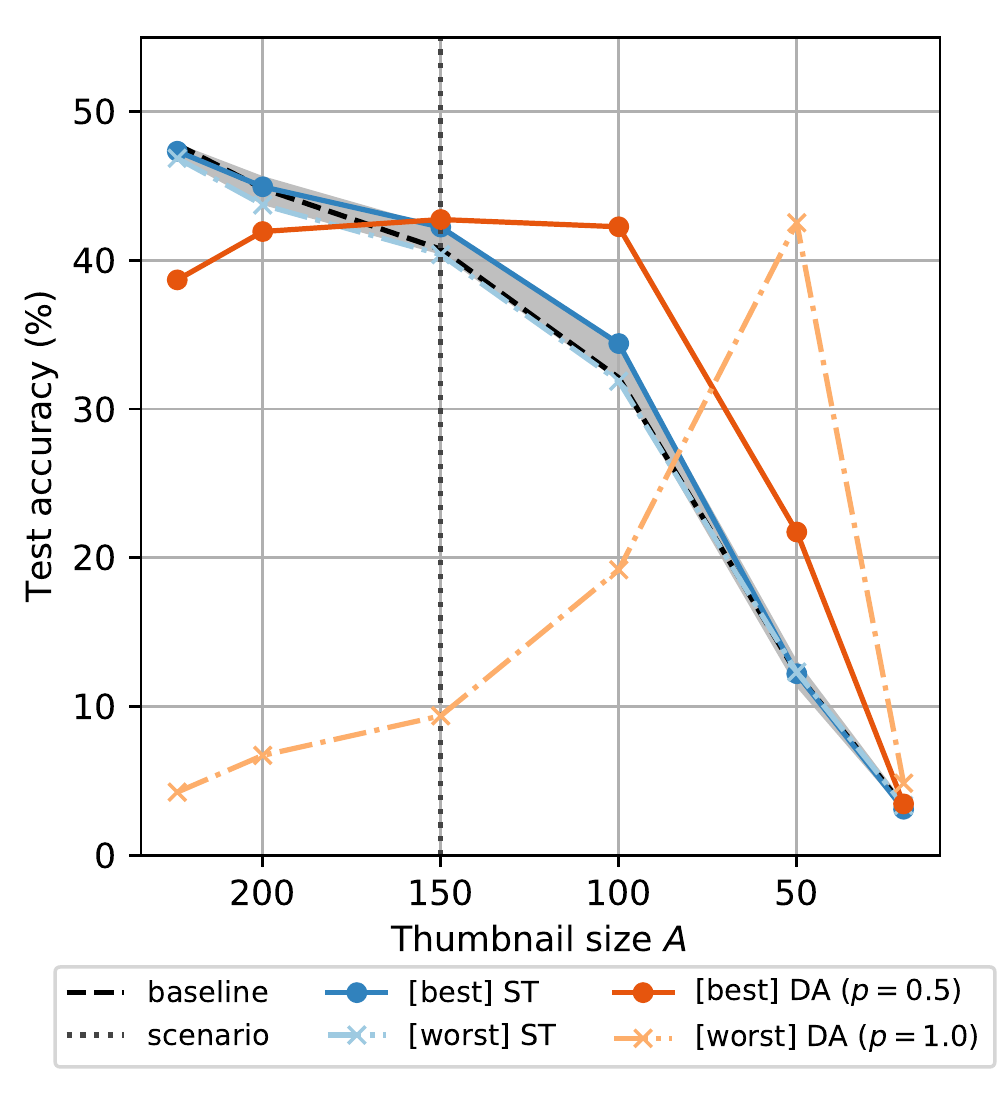}
	\caption{Thumb $\to$ Thumb}
  \end{subfigure}\hfil
  \begin{subfigure}{0.4\textwidth}
	\includegraphics[width=\linewidth]{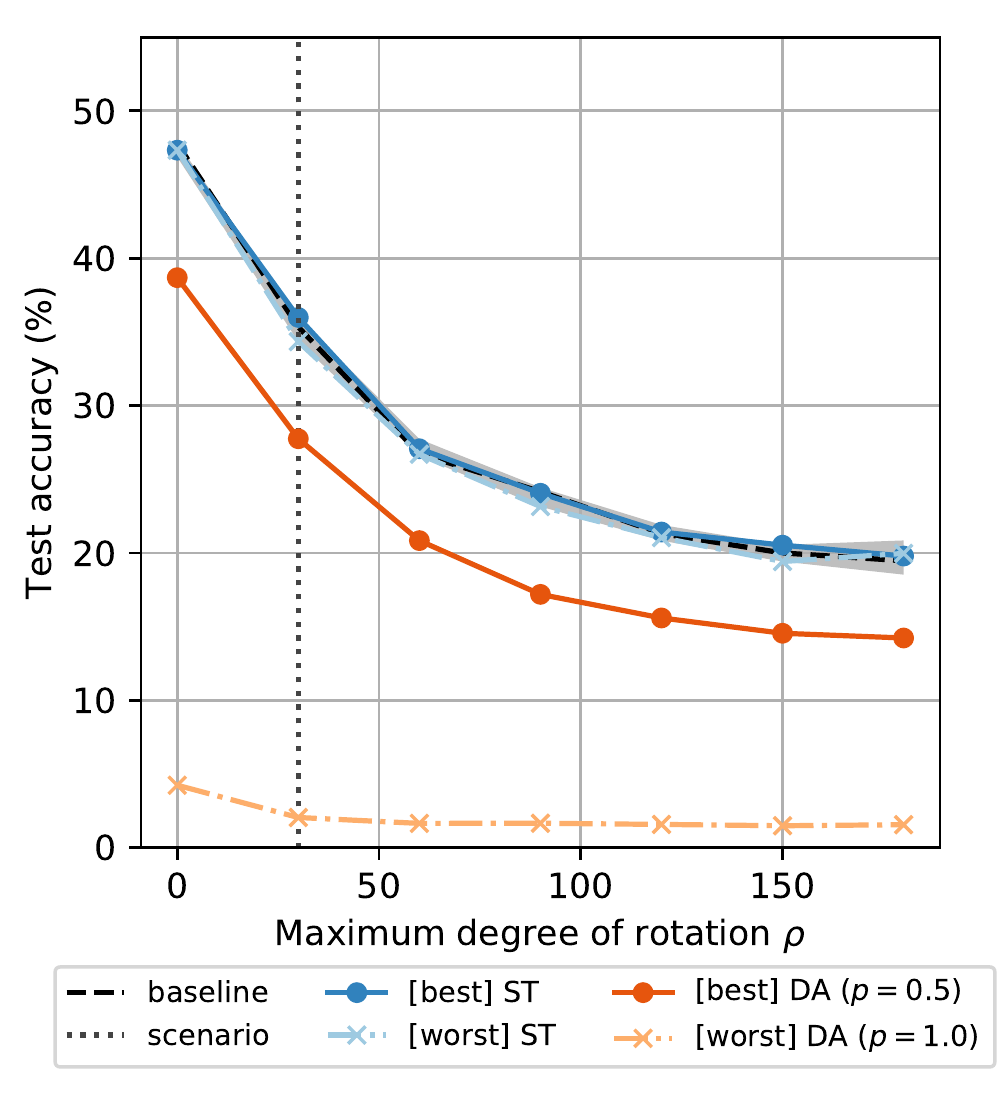}
	\caption{Thumb $\to$ Rotation}
  \end{subfigure}
  \begin{subfigure}{0.4\textwidth}
	\includegraphics[width=\linewidth]{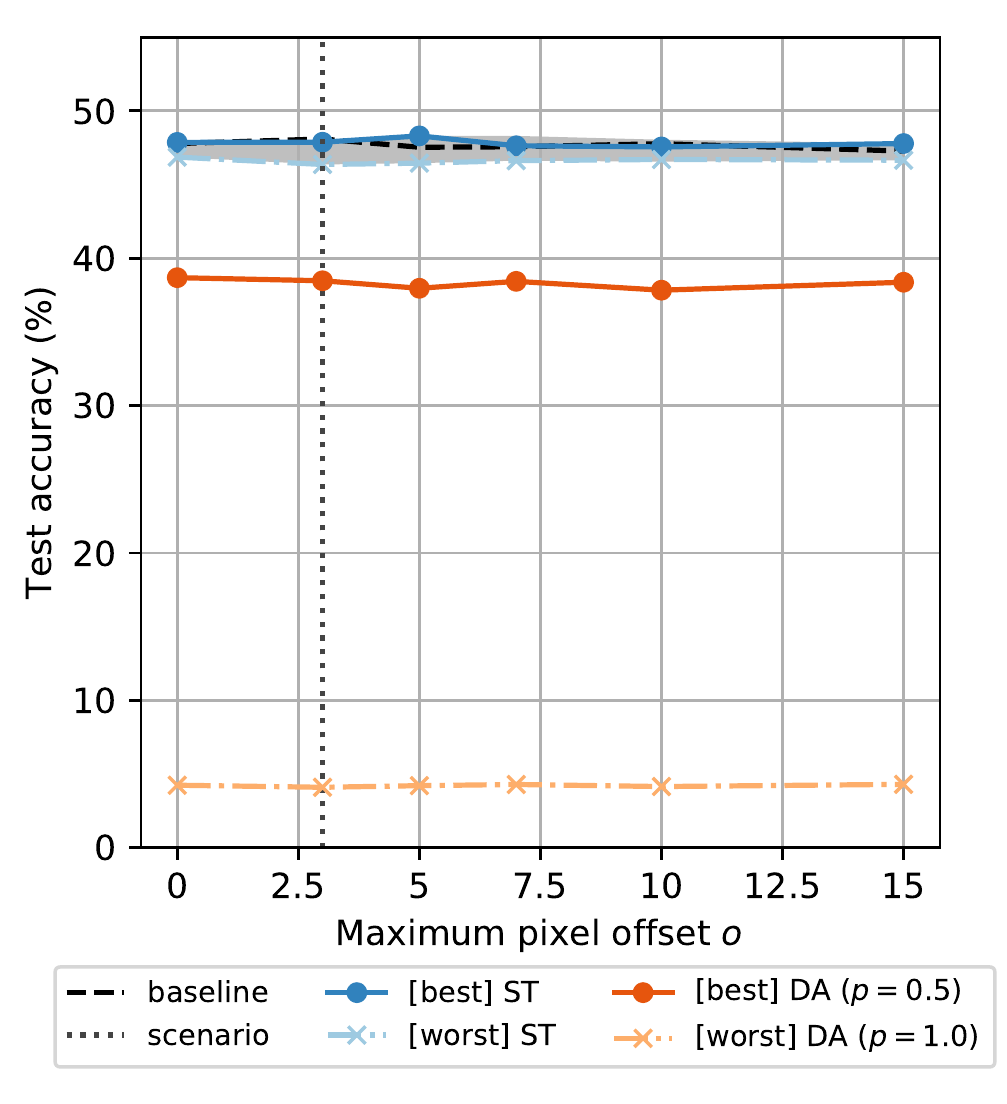}
	\caption{Thumb $\to$ Crops}
  \end{subfigure}\hfil
\caption{Thumbnail resizing as training distorsion.}
\end{figure}

\begin{figure}[htb]
    \centering
\begin{subfigure}{0.4\textwidth}
  \includegraphics[width=\linewidth]{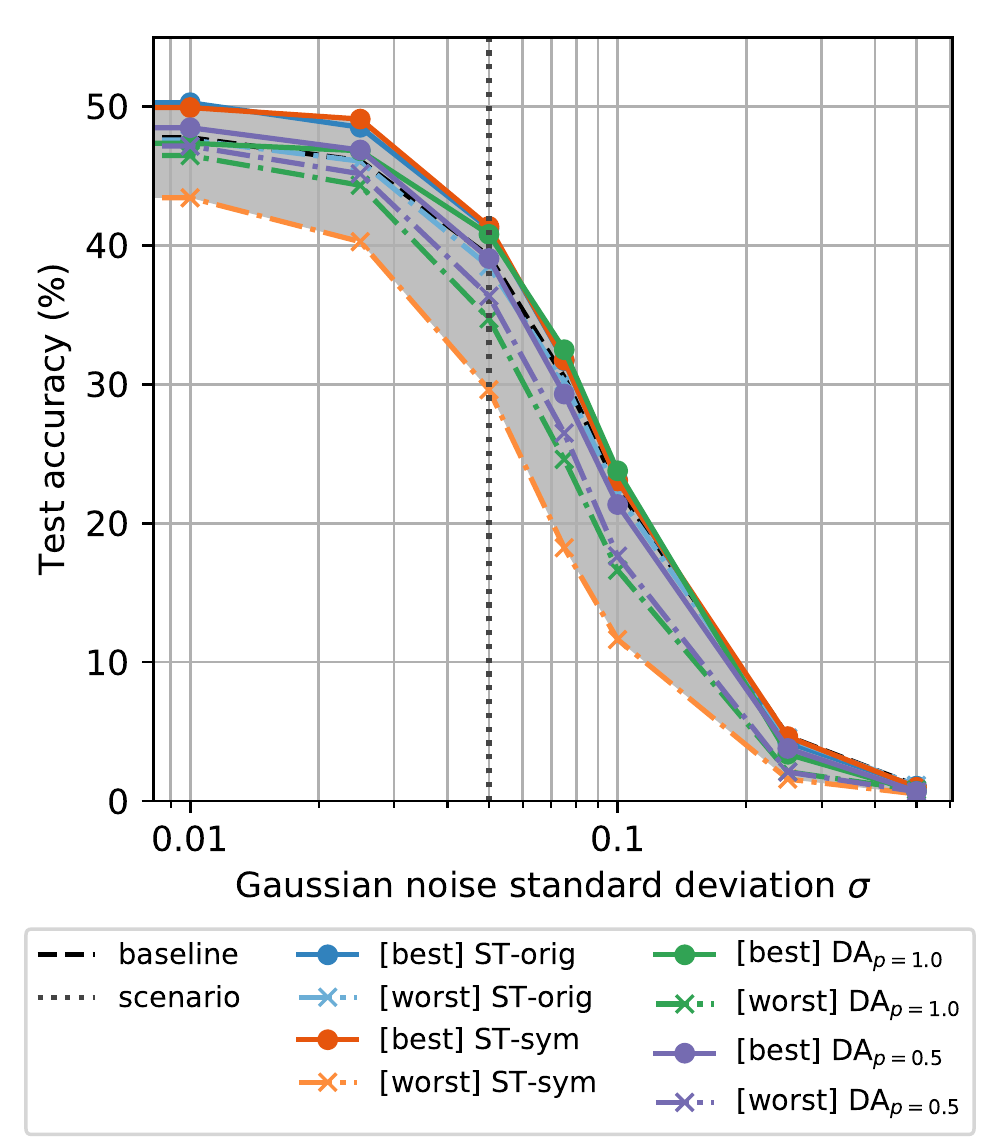}
  \caption{Rotation $\to$ Gauss}
\end{subfigure}\hfil
\begin{subfigure}{0.4\textwidth}
  \includegraphics[width=\linewidth]{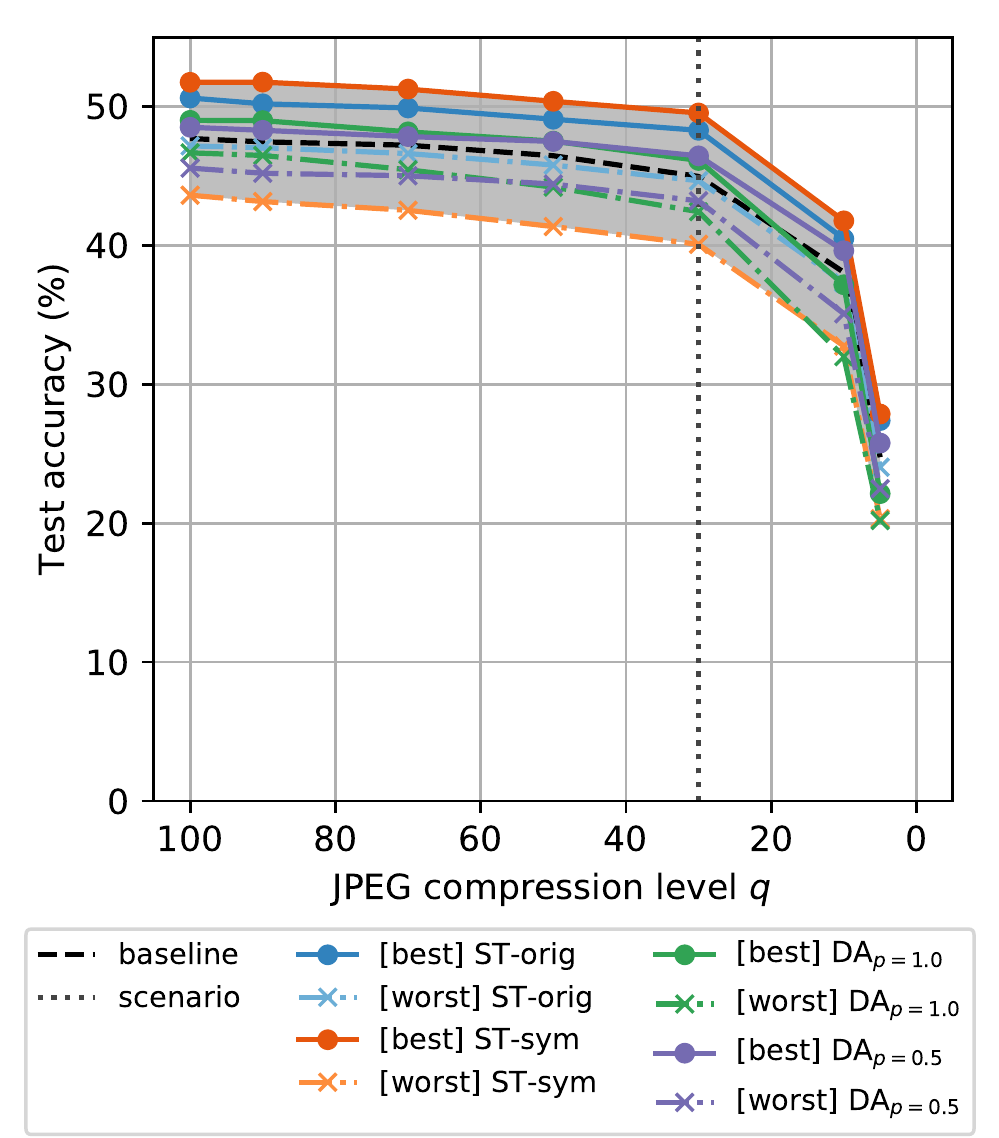}
  \caption{Rotation $\to$ JPEG}
\end{subfigure} 
\begin{subfigure}{0.4\textwidth}
	\includegraphics[width=\linewidth]{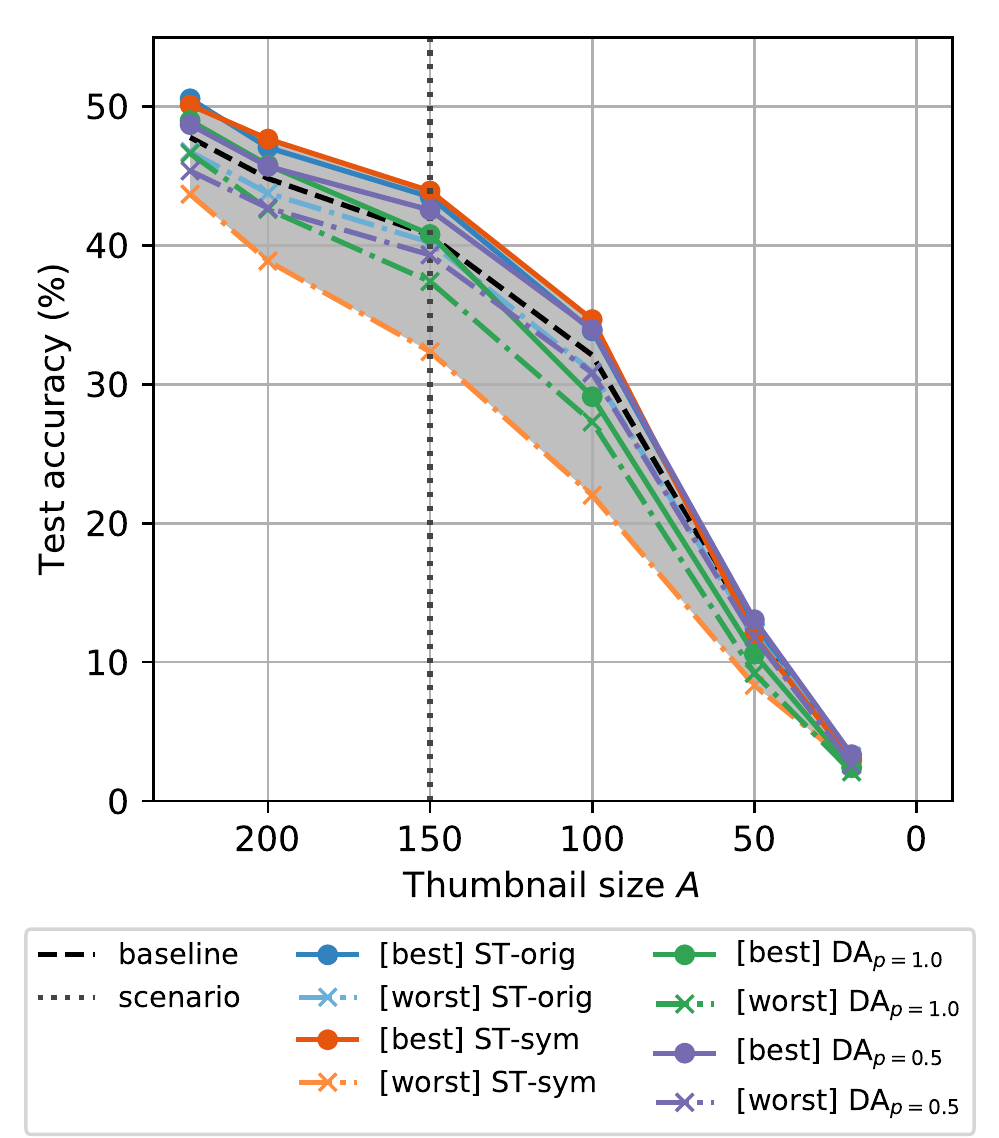}
	\caption{Rotation $\to$ Thumb}
  \end{subfigure}\hfil
  \begin{subfigure}{0.4\textwidth}
	\includegraphics[width=\linewidth]{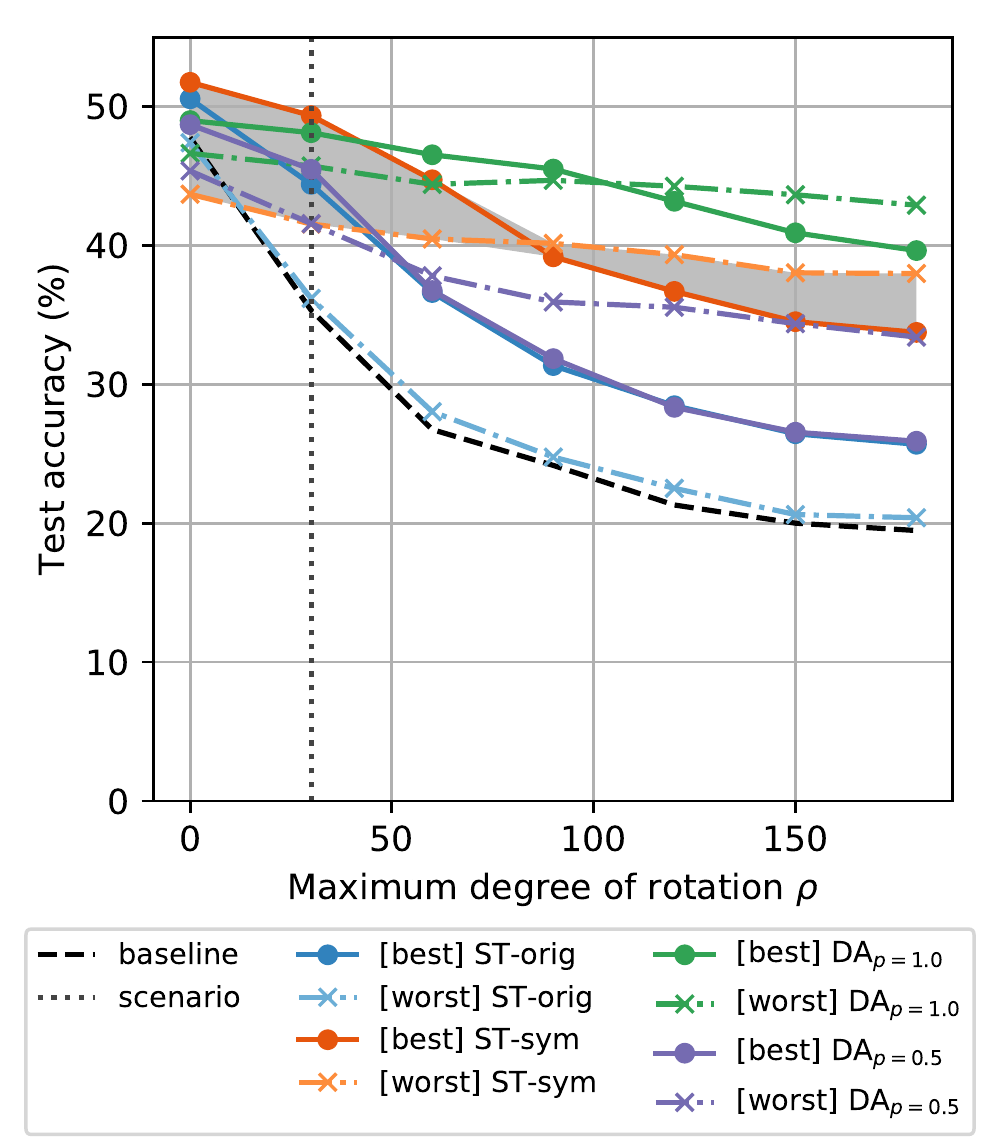}
	\caption{Rotation $\to$ Rotation}
  \end{subfigure}
  \begin{subfigure}{0.4\textwidth}
	\includegraphics[width=\linewidth]{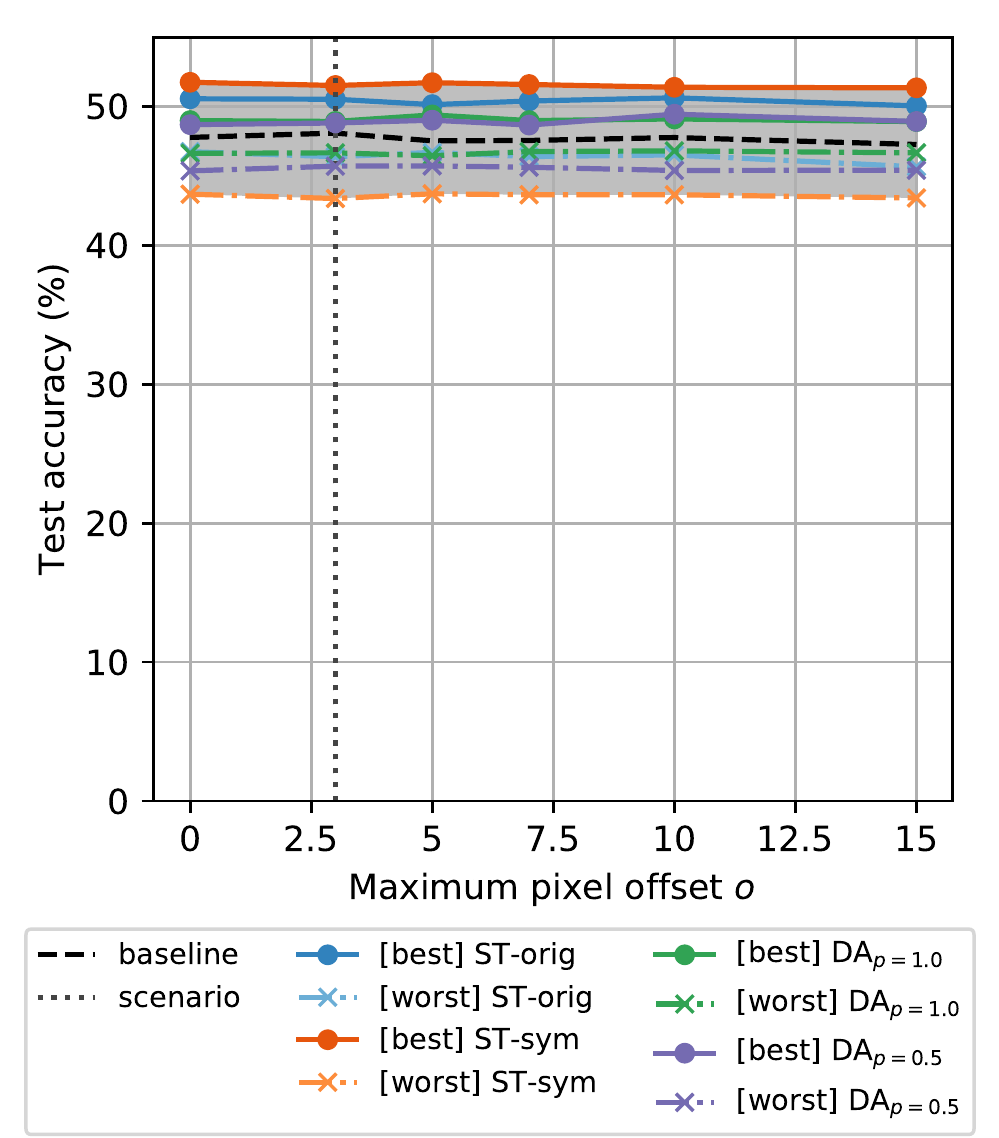}
	\caption{Rotation $\to$ Crops}
  \end{subfigure}\hfil
\caption{Rotation as training distorsion.}
\end{figure}

\begin{figure}[htb]
    \centering
  \begin{subfigure}{0.45\textwidth}
	\includegraphics[width=\linewidth]{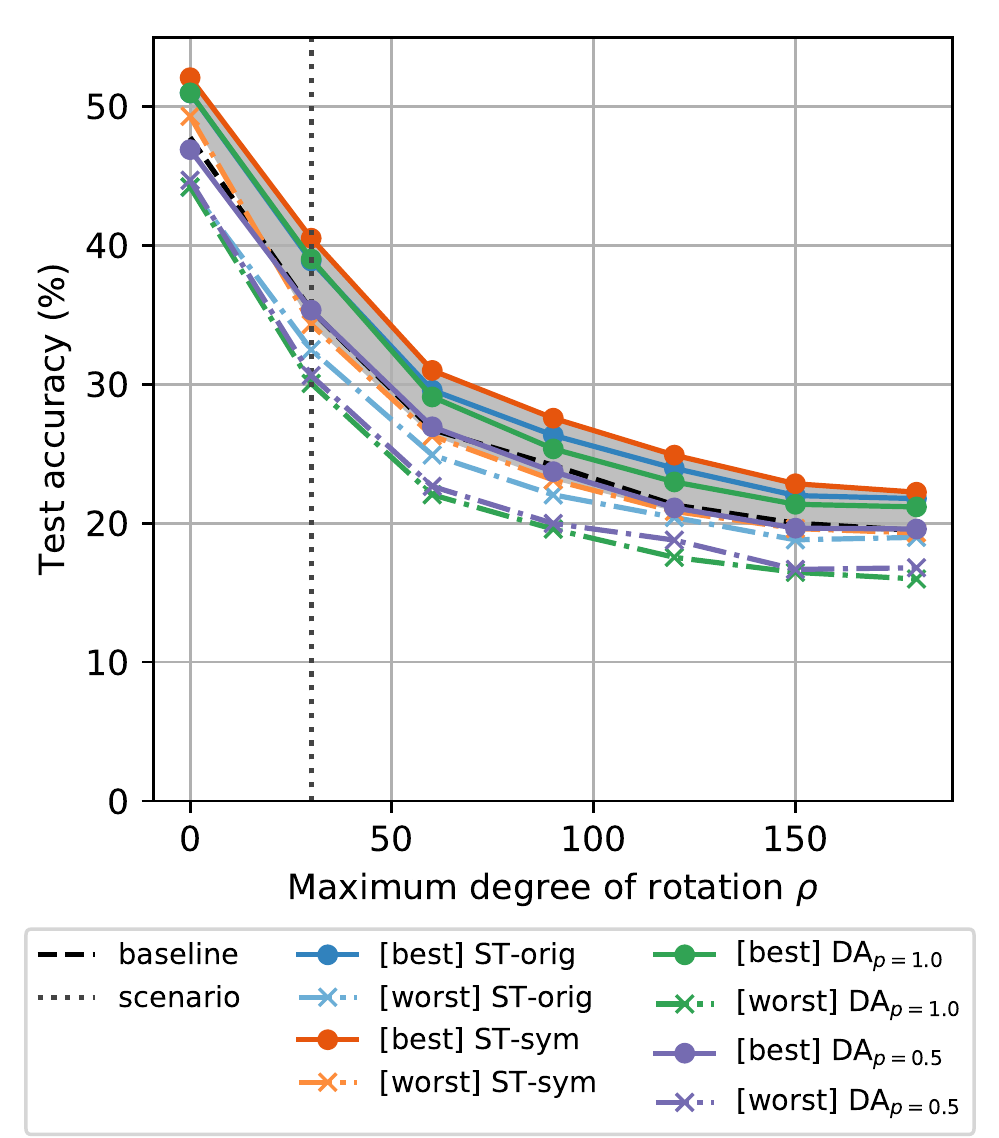}
	\caption{Crops $\to$ Rotation}
  \end{subfigure}\hfil
  \begin{subfigure}{0.45\textwidth}
	\includegraphics[width=\linewidth]{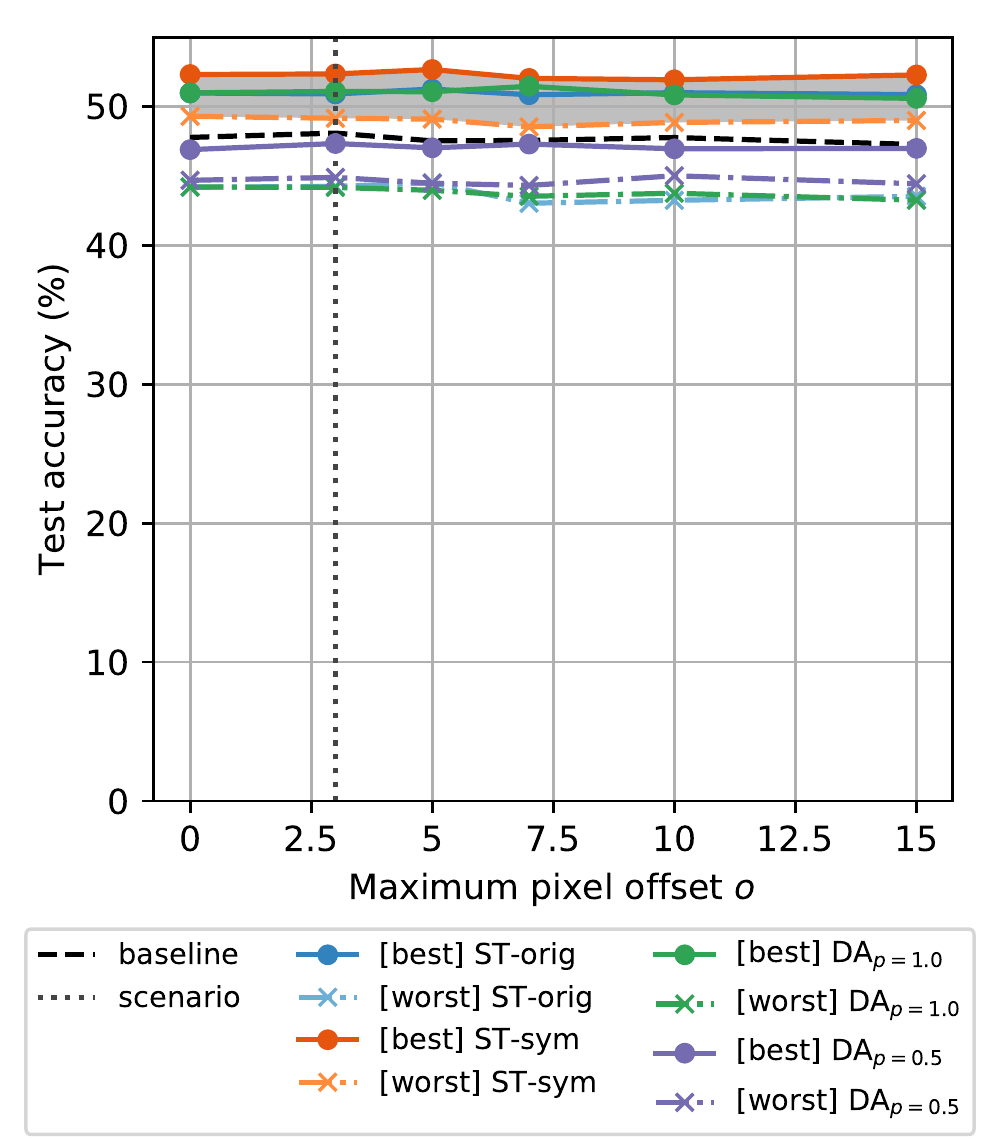}
	\caption{Crops $\to$ Crops}
  \end{subfigure}\hfil
\caption{Crops as training distorsion.}
\end{figure}

\begin{figure}[htb]
    \centering
  \begin{subfigure}{0.45\textwidth}
	\includegraphics[width=\linewidth]{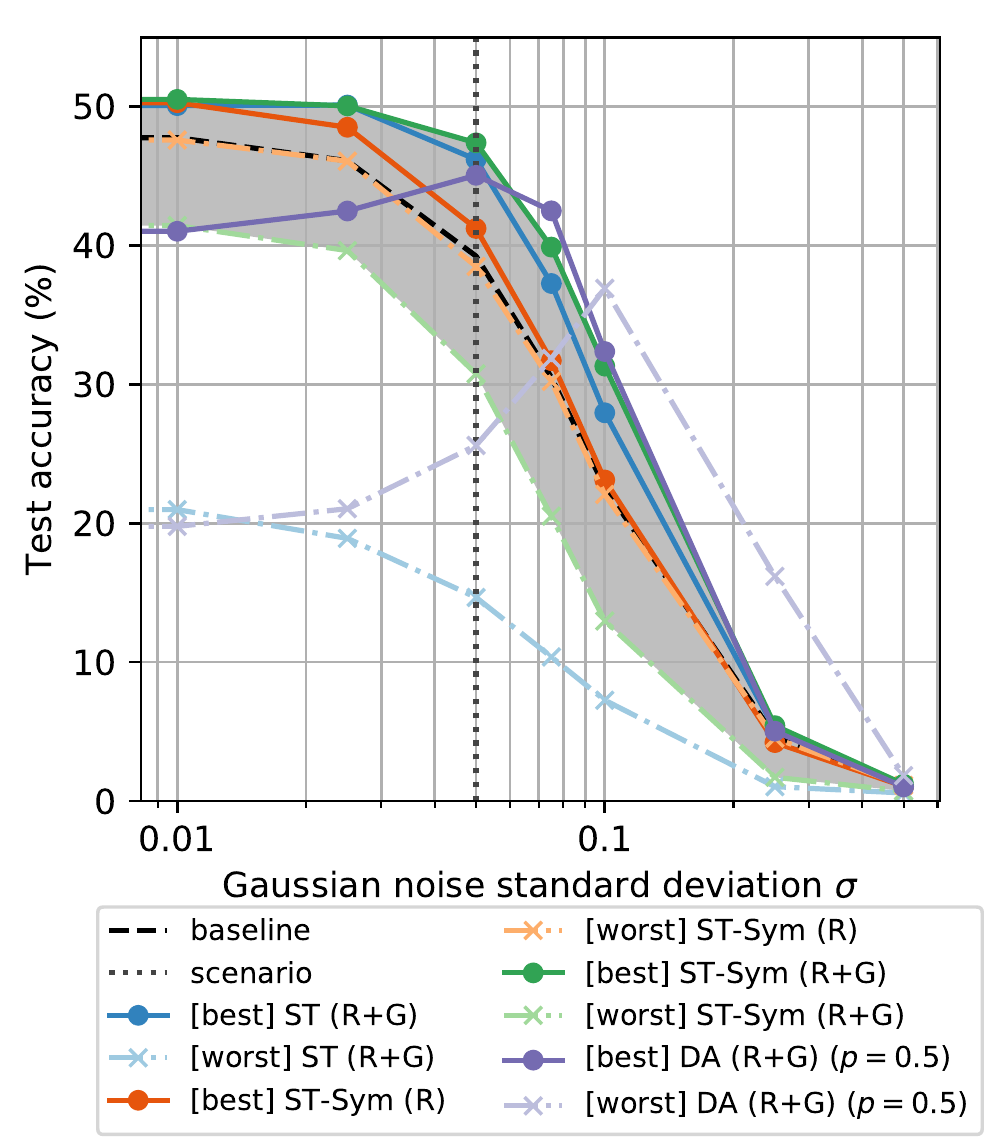}
	\caption{Gauss + Rotation $\to$ Gauss}
  \end{subfigure}\hfil
  \begin{subfigure}{0.45\textwidth}
	\includegraphics[width=\linewidth]{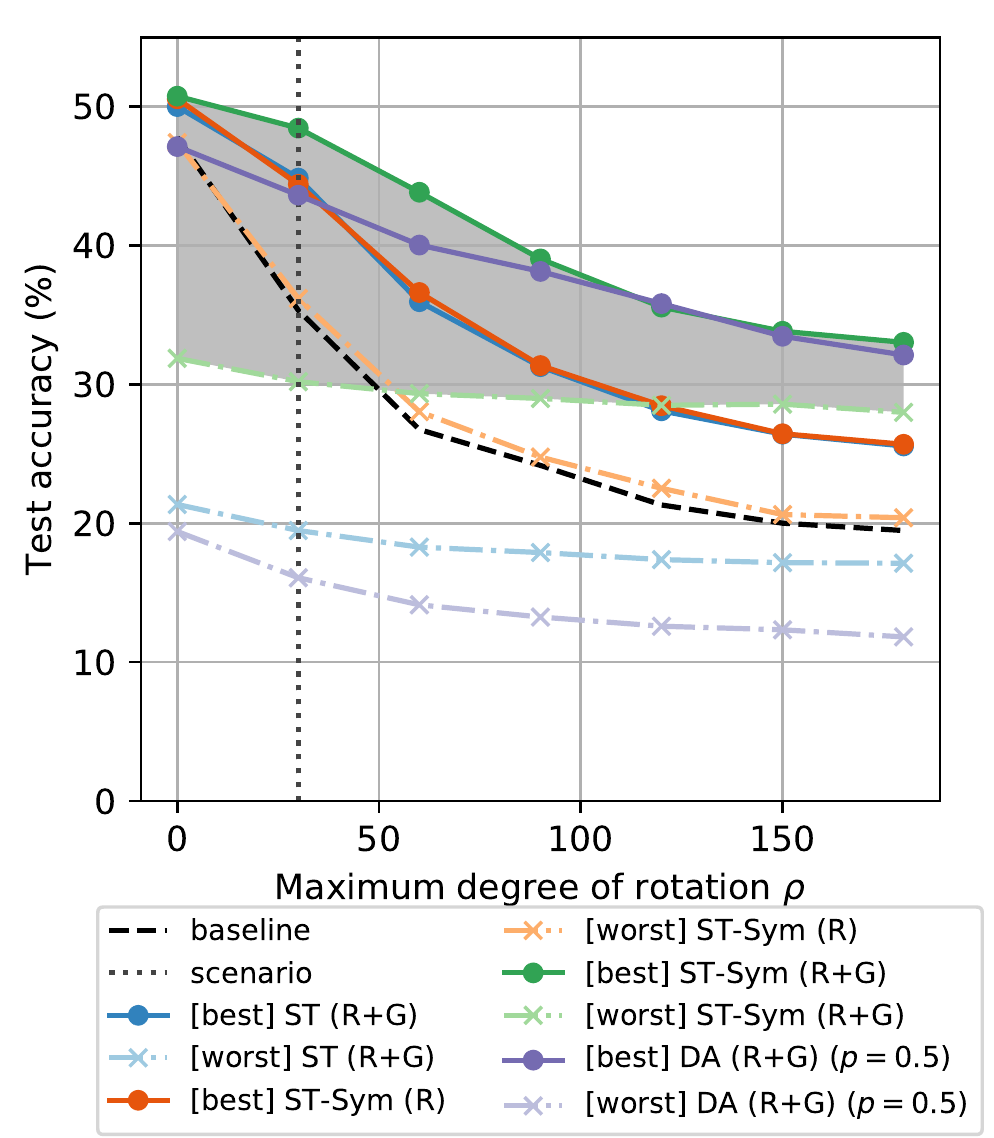}
	\caption{Gauss + Rotation $\to$ Rotation}
  \end{subfigure}\hfil
\caption{Gaussian noise and rotations as training distorsion.}
\end{figure}

\end{document}